\pdfoutput=1
\documentclass[11pt]{article}
\PassOptionsToPackage{hyphens}{url}

\usepackage[preprint]{acl}

\usepackage{times}
\usepackage{latexsym}

\usepackage[T1]{fontenc}
\usepackage[utf8]{inputenc}

\usepackage{microtype}
\usepackage{inconsolata}

\usepackage{graphicx}
\graphicspath{{figures/}}
\usepackage{booktabs}
\usepackage{multirow}
\usepackage{amsmath}
\usepackage{amssymb}
\usepackage{amsfonts}

\title{Comparative Evaluation of LLM-Based Review Paradigms \\ for Academic Paper Quality Assessment}

\author{
 \textbf{Fan Huang\textsuperscript{1}},
 \textbf{Hongyu Zhou\textsuperscript{2,3}}
\\
 \textsuperscript{1}Indiana University Bloomington, Bloomington, Indiana, USA
\\
 \textsuperscript{2}School of Information Resource Management, Renmin University of China, Beijing, China
\\
 \textsuperscript{3}Faculty of Social Sciences, University of Antwerp, Antwerp, Belgium
\\
 \small{\texttt{huangfan@acm.org}, \texttt{hzhou@ruc.edu.cn}}
}

\begin{document}
\maketitle

\begin{abstract}
Growing submission volumes have renewed interest in using large language models (LLMs) to support peer review. It remains unclear how such evaluation should be structured, how much the model matters, and whether LLM scores predict research impact. We evaluate five paradigms across three input representations and three LLMs on 232 stratified ICLR 2025 submissions. Direct absolute rating matches comparative ranking on rank correlation with human scores, yet compresses most predictions above the accept threshold, flagging almost none of the papers reviewers scored below it. A base-rate-matched recalibration control locates the cause: comparative ranking supplies placement, not better ordering, and after recalibration Direct Rating reaches higher agreement than Elo while making 24\% as many API calls. Abstract-only input beats full-paper summarization, and model choice matters as much as paradigm choice. Replicating on all 11{,}513 scored submissions reproduces these patterns, shows absolute agreement depends on the evaluated set's score spread, and finds none beating an always-reject baseline in accuracy. Human scores track citations better than any paradigm, with one exception: Group Rating retains its citation correlation after controlling for human scores, and alone ranks within the accepted pool. Ranking quality alone cannot assess review quality; calibration, thresholds, model, and target outcome all matter.
\end{abstract}

\section{Introduction}
\label{sec:introduction}

\begin{figure*}[t]
\centering
\includegraphics[width=0.8\textwidth]{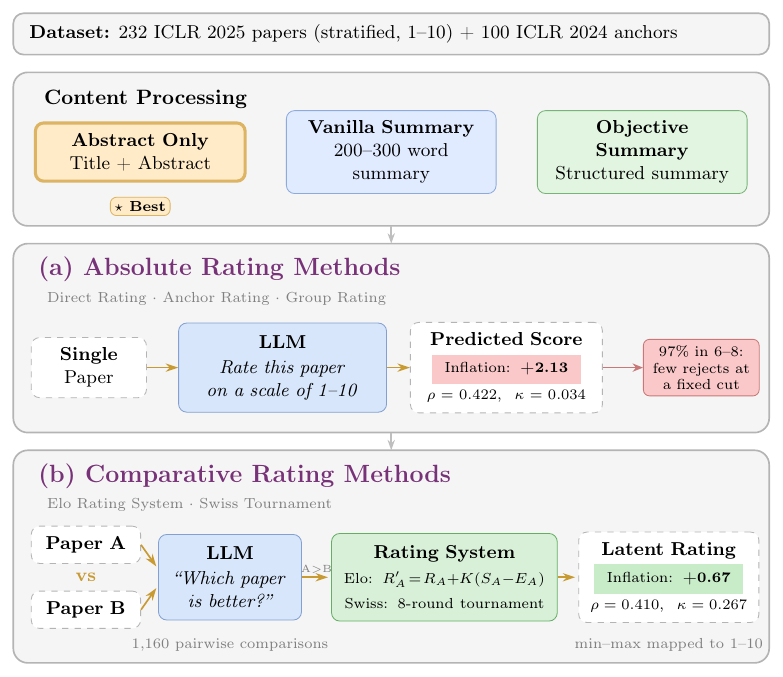}
\caption{Overview of the experimental pipeline. Papers pass through three input representations (top), then through (a) absolute rating methods, which score each paper independently, or (b) comparative rating methods, which infer a latent rating from head-to-head outcomes and rescale it to 1--10. Only the comparative family is rescaled, so inflation and other absolute-error metrics are not comparable across families. Inline metrics give the best configuration in each family on the stratified 232-paper sample at the fixed $\geq 6$ threshold.}
\label{fig:overview}
\end{figure*}

Major machine learning conferences now receive more than 10{,}000 submissions a year without a matching growth in reviewers, renewing concern about the consistency, quality, and sustainability of peer review \citep{checco2021ai,shah2022challenges,wei2025aiimperative,thakkar2025feedback}. Venues have responded at scale: a randomized LLM-feedback trial at ICLR 2025 \citep{thakkar2025feedback}, an AI review on every main-track submission at AAAI 2026 \citep{biswas2026aaai}. LLM-written reviews also appear unprompted, motivating detection work \citep{yu2026detection,rao2025detecting}. Large language models (LLMs) read, compare, and evaluate text competently \citep{openai2023gpt4,zheng2023judging}, and a growing literature asks how much of this load they can absorb \citep{zhuang2025survey}.

Using an LLM to assess a paper is not one method but a design space with three largely independent choices: the \emph{review paradigm}, the procedure that turns a paper into a score; the \emph{input representation}, how much of the paper the model reads; and the model itself. Absolute scoring, the most immediate paradigm, is hard to calibrate: LLM judges show systematic scoring biases, and single-answer scores are argued to be less stable than pairwise preferences \citep{zheng2023judging,wang2024fair}; their scores favor specific scale values \citep{li2025scoringbias,fujinuma2026contrastive} and, for most models, exceed human ratings \citep{li2026llmreviewer}. We call this \emph{grade inflation}, the mean signed difference between predicted and human scores. Since an accept/reject decision is a threshold on a score, inflation can leave a system that orders papers sensibly unable to decide about any.

Prior work predicts acceptance from manuscript features \citep{kang2018dataset}, generates review text \citep{yuan2022can,gao2024reviewer2,darcy2024marg}, and studies frontier models as reviewers \citep{liang2023gpt4,robertson2023gpt4,liu2023reviewergpt,lu2024aiscientist}, while a position paper argues deployment is outrunning its evaluation \citep{baumann2026stop}. Remedies for miscalibration include calibrating the judge's scoring criteria against human labels \citep{liu2023calibrating} and comparative evaluation, which infers quality from head-to-head outcomes rather than an absolute scale: Elo \citep{elo1978rating} and Swiss tournaments \citep{olafsson1990weighted} are self-calibrating in this sense, and Elo is increasingly adopted for LLM evaluation \citep{boubdir2023elo}, and pairwise comparison has been applied to peer review \citep{zhang2025replication}.

Two gaps remain. Existing comparisons vary the paradigm with the model largely fixed \citep{zhang2025replication} or compare models under one protocol \citep{lu2024aiscientist,li2026llmreviewer}, so it is unclear which choice dominates the outcome or whether comparative ranking helps for paper assessment specifically. And agreement with human scores is often checked by correlation \citep{lu2024aiscientist,zhang2025replication}; where accept/reject decisions are evaluated \citep{lu2024aiscientist}, the link between a high correlation and useful decisions, and its dependence on the evaluation sample, is not tested.

We therefore ask three questions. RQ1, paradigm design: how should an LLM evaluate papers, and do abstracts alone carry sufficient signal? We compare absolute paradigms (Direct, Anchor, Group Rating) against comparative ones (Elo, Swiss tournament) under three input representations (Figure~\ref{fig:overview}). RQ2, model dependence: does model choice matter more than paradigm choice? We replicate every paradigm across three LLMs. RQ3, external validity: do predicted scores relate to research impact? We correlate them with citation counts.

We separate \emph{ranking potential}, how a method orders papers, from \emph{decision utility}, the accept/reject decisions it induces, and find they diverge: Elo and Direct Rating track human scores comparably ($\rho = 0.410$ vs.\ $0.422$), yet Direct flags 4.3\% of below-threshold papers against Elo's 66.4\%. LLM reviewers that accept nearly every paper have been reported before \citep{verma2026allaccept}, as have LLMs that, asked to verify an answer, accept a wrong one at least 76\% of the time \citep{huang2026reflect}; we term the paradigm-dependent, fixed-threshold form of this \emph{reject blindness}. A recalibration control traces this to placement rather than ordering: at its own base rate Direct overtakes Elo while making 24\% as many API calls. Abstract-only input beats full-paper summarization, model choice matters as much as paradigm choice, agreement depends on the evaluated set's score distribution, and human scores track citations better than any paradigm tested.

\section{Related Work}
\label{sec:related}

\subsection{Automated and LLM-Assisted Peer Review}
Early work framed peer review as prediction over manuscript features (the PeerRead dataset; \citealp{kang2018dataset}) or as learned review generation \citep{yuan2022can}. Instruction-tuned LLMs shifted attention to generating reviews directly: \citet{liang2023gpt4} found partial overlap between GPT-4 feedback and human reviewers, while smaller studies find GPT-4 reviews comparable to human ones in average helpfulness but more variable \citep{robertson2023gpt4}, and GPT-4 useful for targeted checks such as finding inserted errors and verifying checklist items, though not yet as an evaluator of whole papers \citep{liu2023reviewergpt}. Later systems add generated aspect prompts \citep{gao2024reviewer2}, multi-agent collaboration \citep{darcy2024marg}, or full discovery pipelines \citep{lu2024aiscientist}; \citet{zhuang2025survey} survey the area. Deployment has since moved ahead of this literature \citep{thakkar2025feedback,biswas2026aaai}: across 75{,}800 reviews of 19{,}490 ICLR 2026 submissions \citet{baumann2026stop} report that AI reviewers agree with one another excessively, and that on 60 randomly sampled papers an automated, mostly stylistic rewrite with no new experiments raised their scores, and a benchmark of twelve models finds that most rate papers higher than human reviewers do \citep{li2026llmreviewer}.

\subsection{Reliability and Bias of LLM Judges}
LLM judges are widely used for open-ended evaluation \citep{zheng2023judging} but show position and order effects \citep{wang2024fair,huang2026cogbias}, rubric- and score-related biases \citep{li2025scoringbias}, and sensitivity to the rating scale itself \citep{fujinuma2026contrastive,huang2024ratesscales}. Repeated runs of one prompt also yield materially different scores \citep{haldar2025rating}, agreement can look strong without validity following \citep{norman2026reliability}, and reported agreement depends on protocol choices made before any statistic is computed \citep{rao2026agreement}. These motivate calibrating the judge's scoring criteria against human labels \citep{liu2023calibrating} and comparative paradigms that avoid an absolute scale, though comparative protocols carry biases of their own \citep{tripathi2025pairwise}.

\subsection{Comparative and Ranking-Based Evaluation}
Elo \citep{elo1978rating} and Swiss tournaments \citep{olafsson1990weighted} estimate latent quality from pairwise outcomes, and Elo is increasingly adopted in LLM evaluation \citep{boubdir2023elo}. Their assumptions are not innocuous: LLM judges hold non-transitive preferences \citep{xu2025nontransitivity}, and Elo's sequential update makes ratings depend on the replay order \citep{liu2025amelo}, which is why we fix the pairing schedule in advance, although our ratings are still updated in the order in which the comparisons were returned. Closest to our work, \citet{zhang2025replication} report that pairwise comparison selects more-cited papers than absolute rating, whose scores mostly fell at 6 and 7; against human scores, our recalibration control locates the comparative advantage in score placement rather than ordering, since at each paradigm's own base rate it disappears. \citet{epstein2026allocate} aggregate LLM comparisons under a Bradley--Terry model to route marginal reviews; \citet{tong2026taste} instead train a judge against citation outcomes, while a zero-shot GPT-4o-mini rater of disruption potential, reading only titles and abstracts, identifies the 2\% of ICLR papers most cited within ICLR below chance (AUC 0.424; \citealp{huang2026catalyst}).

\section{Methodology}
\label{sec:methodology}

\subsection{Dataset Construction}
\label{sec:dataset}

\paragraph{Test set (ICLR 2025).}
We constructed a test set of 232 ICLR 2025 submissions, stratified to 30 papers in each score bin from 2 to 8 and every available paper in the sparse bins 1, 9 and 10 (7, 13 and 2 papers). Ground-truth scores are the mean OpenReview reviewer ratings. That target is itself noisy: over the 8{,}655 non-withdrawn ICLR 2025 submissions, a one-way random-effects analysis \citep{shrout1979intraclass} of the per-reviewer scores gives ICC(1,1) $= 0.327$ for a single reviewer and ICC(1,$k$) $= 0.663$ for the mean of the 4.05 reviewers a paper receives on average. Correcting for attenuation gives a ceiling of $\rho = 0.815$ here and, on the wider-spread stratified set, $\rho = 0.951$; correlations should be read against the ceiling of their own sample rather than against 1. The resulting distribution has a mean of 5.16 and a standard deviation of 2.16, spanning strong reject (1.0) to strong accept (10.0).

\paragraph{Anchor set (ICLR 2024).}
For calibration-based methods we built an anchor set of 100 ICLR 2024 papers with zero score variance. Unanimous papers are stable calibration points, but selecting on zero variance collapses their range: standard deviation 1.66 around a mean of 5.14, never above 8.0, against 2.16 and the full 1--10 range in the test set, and narrower still (0.93) once averaged over the three anchors retrieved per paper. The calibration points therefore span little of the scale they are meant to calibrate, which we take to be the main reason Anchor Rating performs poorly (Appendix~\ref{app:resolution}). Details of the PDF-to-text preprocessing pipeline are given in Appendix~\ref{app:preprocessing}.

\subsection{Review Paradigms}
\label{sec:paradigms}

We evaluate five paradigms.

\emph{Direct Absolute Rating} (hereafter \emph{Direct}) asks the LLM to rate each paper independently on a 1--10 scale, using the explicit score bands and criteria given in Appendix~\ref{app:prompts}. It uses one call per paper and is susceptible to grade inflation.

\emph{Anchor Rating} retrieves the three most similar anchor papers by cosine similarity of abstract embeddings (text-embedding-3-small) and presents them with their known scores as calibration points. The same device, the most similar example shown with its scores, raised ChatGPT's agreement with human judgments of explanation quality in paired comparisons but lowered it on a 7-point scale \citep{huang2024ratesscales}.

\emph{Group Rating} presents papers in sets of five and asks the LLM to score all five jointly on a 1--10 scale, yielding 46 groups with each paper evaluated once. Groups and within-prompt order are randomized; an earlier grouping ordered by score depressed the paradigm, and a prompt-position effect survives randomization (Appendix~\ref{app:paradigms}). It remains absolute: the group supplies context, but the output is still a score on a fixed scale.

Under \emph{Elo Rating} the LLM is asked, for each binary comparison, which paper is better; ratings update by the Elo rule \citep{elo1978rating} in its common logistic form
\begin{align}
    E_A &= \frac{1}{1 + 10^{(R_B - R_A)/s}}, \label{eq:expected}\\
    R'_A &= R_A + K\,(S_A - E_A), \label{eq:update}
\end{align}
where $E_A$ is the expected score for $A$ against $B$, $R_A$ and $R_B$ their current ratings, $S_A \in \{0, \tfrac{1}{2}, 1\}$ a loss, tie or win, and $s = 400$, $K = 32$ the conventional scale and update step. Each paper receives 10 comparisons on average on the 232-paper set (1{,}160 total), with final ratings min--max normalized to 1--10.

\emph{Swiss Tournament} runs $\lceil\log_2 n\rceil$ rounds (8 here): round 1 is random, later rounds pair papers of similar current score while avoiding repeats, and win/tie/loss records are normalized to 1--10 the same way (Appendix~\ref{app:paradigms}).

\subsection{Input Representations and Implementation}

We tested three input representations: \emph{Abstract-Only} (title and abstract only), \emph{Vanilla Summarization} (an LLM-generated 200--300 word summary of the full paper), and \emph{Objective Summarization} (the same length under Background, Method, Results and Limitations headings). Appendix~\ref{app:prompts} gives the prompt templates.

Unless otherwise noted, the 232-paper experiments used GPT-4o-mini \citep{openai2024gpt4omini} at temperature 0.3 with 200--500 output tokens depending on the task, and text-embedding-3-small for anchor similarity. The cross-model replication used the same temperature, the same Direct prompt, condensed versions of the Anchor, Group and binary-comparison prompts (the same tasks and JSON output, with the criteria on one line, the 1--10 score bands of Anchor and Group on one line without their descriptions, and a shorter reasoning instruction for Group), and the same token budgets with one exception recorded in Appendix~\ref{app:paradigms}, serving \texttt{meta-llama/Llama-3.3-70B-Instruct-Turbo} \citep{grattafiori2024llama3} and \texttt{openai/gpt-oss-120b} \citep{openai2025gptoss} through the Together AI API. The 232-paper design comprised 8{,}258 GPT-4o-mini calls, 7{,}794 for evaluation across the fifteen paradigm-by-representation cells and 464 for summarization; population-scale cost is reported in Appendix~\ref{app:cost}.

\subsection{Evaluation Metrics}

\paragraph{Correlation and regression.}
We report Pearson's $r$, Spearman's $\rho$, and Kendall's $\tau$, together with Mean Squared Error (MSE). Grade inflation is the mean signed difference between predicted and ground-truth scores, so positive values indicate systematic over-prediction. One asymmetry matters. Direct, Anchor and Group Rating emit 1--10 scores directly, whereas Elo and Swiss produce latent ratings min--max mapped onto that scale (Section~\ref{sec:paradigms}); fixing the endpoints at 1 and 10 pulls their predictions toward the middle of a scale whose ground-truth mean is 5.16, independently of judgment quality. Inflation is therefore not like-for-like across families, and neither is any metric on absolute distances, including MSE. Rank correlations and threshold-free measures are unaffected, and Section~\ref{sec:recalibration} separates placement from ordering.

\paragraph{Classification.}
For binary accept/reject decisions (Accept if the score is $\geq 6$, applied alike to ground-truth and predicted scores) we report accuracy, AUC-ROC on the continuous predictions, Cohen's $\kappa$, and macro and weighted F1. We refer to $\kappa$ and macro F1 jointly as \emph{decision utility}: the quality of the accept/reject decisions a system induces, as distinct from the quality of the ordering it produces. These are our primary utility metrics, since weighted F1 misleads under class imbalance: a system that accepts everything still attains a non-trivial weighted F1 but zero \emph{reject recall}, the fraction of papers human reviewers scored below the threshold that the system also rejects.

\paragraph{Statistical tests.}
We use bootstrap confidence intervals (1{,}000 resamples unless otherwise noted) for correlations, paired bootstrap tests for paradigm comparisons, and Wilcoxon signed-rank tests for absolute-error differences.
\section{RQ1: Paradigm Design}
\label{sec:results}

How should an LLM evaluate papers, and do abstracts alone carry sufficient signal? We take these in turn: input representation, then the five paradigms, their calibration and reliability, the accept/reject decisions they induce, a population-scale replication, and a recalibration control that asks which of these effects is about ordering and which about score placement.

\subsection{Input Representation and Paradigm Comparison}
\label{sec:summarization}

We first isolate the effect of input representation by averaging metrics across all five paradigms (Appendix~\ref{app:full}).

Abstract-only evaluation outperforms both summarization methods on every correlation metric ($r = 0.333$ against 0.193; $\rho = 0.334$ against 0.180), and the ordering holds within all five paradigms. Because our summarization pipeline is lossy (Appendix~\ref{app:preprocessing}), this compares abstract-only input against a summary rather than against full-paper reading.

We therefore compare the five paradigms under abstract-only evaluation (Table~\ref{tab:paradigm}).

\begin{table}[htbp]
\centering
\small
\begin{tabular}{@{}lccccc@{}}
\toprule
\textbf{Paradigm} & \textbf{$r$} & \textbf{$\rho$} & \textbf{$\tau$} & \textbf{MSE} & \textbf{Infl.} \\
\midrule
Direct & \textbf{0.405} & \textbf{0.422} & \textbf{0.351} & 8.42 & +2.13 \\
Elo & 0.401 & 0.410 & 0.285 & \textbf{4.85} & +0.67 \\
Anchor & 0.353 & 0.369 & 0.300 & 10.35 & +2.51 \\
Group & 0.405 & 0.388 & 0.279 & 7.87 & +1.98 \\
Swiss & 0.100 & 0.080 & 0.057 & 6.71 & \textbf{+0.04} \\
\bottomrule
\end{tabular}
\caption{Review paradigm comparison under abstract-only input (GPT-4o-mini). Column headings as in Table~\ref{tab:summarization}. The Group row is a three-run mean (Appendix~\ref{app:paradigms}); all other rows are single runs.}
\label{tab:paradigm}
\end{table}

Direct Rating and Elo achieve comparable correlation with human scores ($\rho = 0.422$ vs.\ $0.410$; paired bootstrap $\Delta\rho = +0.012$, 95\% CI $[-0.127, +0.151]$ from 10{,}000 resamples). Group Rating follows at $\rho = 0.388$; all three outperform Swiss ($p < 0.001$), whose correlation is not significant, so its ranking may not reliably reflect quality. The families differ mainly in where they place scores: Direct and Anchor compress 97\% and 94\% of predictions into the 6--8 range against a ground-truth $\sigma$ of 2.16, whereas the comparative paradigms spread wider, partly through the min--max mapping (Appendices~\ref{app:full} and \ref{app:summdist} give per-paradigm figures and the full 15-combination matrix). Because almost all Direct scores fall above the accept threshold, that paradigm separates few of the papers reviewers scored below it even though its correlation is high; Appendix~\ref{app:ranking} shows that a top-20 cut is not even well defined for it, since 87 of the 232 papers share its highest score.

\subsection{Grade Inflation and Judge Reliability}
\label{sec:reliability}

Grade inflation varies systematically by paradigm type: the comparative methods range from +0.04 to +0.67, the absolute ones from +1.98 to +2.78, and Wilcoxon tests put Elo's absolute prediction errors below every other paradigm (all $p < 0.01$; Appendix~\ref{app:paradigms}). A single Direct rating also carries substantial decoding noise: three runs on the full population give ICC(2,1) $=0.417$ \citep{shrout1979intraclass}, the ``poor'' band of \citet{koo2016guideline}, so single-shot Direct correlations are a lower bound (Appendix~\ref{app:reliability}). Comparative paradigms cannot be scored with ICC(2,1); for Elo we compare ratings at two comparison densities instead (Appendix~\ref{app:resolution}), and we have no such diagnostic for Swiss.

\subsection{Classification Performance}
\label{sec:classification}

Appendix~\ref{app:classification} reports the binary classification metrics, which reveal a divergence between methods with similar correlation.

This subsection and the next apply a single fixed threshold to every paradigm, a choice Section~\ref{sec:recalibration} revisits. At that threshold, correlation alone misleads. Direct Rating reaches an AUC-ROC of 0.682, indistinguishable from Elo's 0.672, so its scores carry discriminative signal; yet inflation shifts the boundary until accuracy is near chance (42.2\% against 60.3\% for always rejecting) and agreement negligible ($\kappa = 0.034$), while Elo reaches 64.2\% accuracy, $\kappa = 0.267$, and 66.4\% reject recall against Direct's 4.3\%. Table~\ref{tab:classification} is computed on the stratified 232-paper sample, whose scores are spread evenly across the quality range; Section~\ref{sec:population} verifies the pattern on the full population, whose scores instead concentrate near the accept/reject boundary.

\subsection{Population-Scale Replication}
\label{sec:population}

Re-running the GPT-4o-mini abstract-only pipeline for Direct, Elo and Swiss on all 11{,}513 scored ICLR 2025 submissions reproduces the pattern and sharpens it: Direct flags $0.5\%$ of below-threshold papers against $4.3\%$ on the stratified sample, Elo's $\kappa$ falls from $0.267$ to $0.076$, and no configuration reaches the $59.5\%$ accuracy of always rejecting. The fall in $\kappa$ is not a prevalence effect, since the accept rate is nearly identical in the two designs; what changes is the spread of ground-truth scores. We therefore recommend reporting absolute $\kappa$ on a \emph{deployment-representative} sample, whose score distribution matches the pool the system would actually screen. Appendix~\ref{app:populationtab} gives the full analysis, the paradigm-specific error trade-offs and Table~\ref{tab:population}.

\subsection{Ranking or Placement? A Recalibration Control}
\label{sec:recalibration}

Applying one fixed threshold to every paradigm does not separate how well a paradigm \emph{orders} papers from where its scores happen to sit. We therefore repeat the population-scale analysis with each paradigm thresholded at its own base-rate-matched cut, the value at which its predicted reject rate equals the true rate of $59.5\%$. The transformation is monotone, so it changes no ranking metric (Table~\ref{tab:population}). It does assume the true reject rate is known, which a deployed system would have to estimate; we use it here as a diagnostic that isolates ordering from placement, not as a deployable procedure.

This qualifies the earlier reading substantially. Direct rises from $\kappa = 0.004$ to $0.112$, ahead of Elo (0.079) and Swiss (0.078); the gap is $+0.032$, 95\% CI $[+0.006, +0.058]$, so unlike the ranking comparison it separates them. Reject blindness is therefore a property of Direct Rating \emph{at a fixed threshold} rather than a defect in its ordering, as its AUC-ROC of 0.682 and population $\rho$ of 0.204 already indicated. What Elo supplies is placement, part of which follows from the min--max mapping rather than the comparison mechanism. Since Direct uses 3 API calls per paper against Elo's 12.5, recalibration is the cheaper route to the better decision quality, so we do not recommend Elo as a default. The two families do make partly independent errors, but averaging their outputs does not reliably exploit this (Appendix~\ref{app:spread}). A single human reviewer scored against the mean of the others reaches $\rho = 0.445$ and $\kappa = 0.341$, so the best paradigm attains under half that ranking. Note also that no configuration reaches the accuracy of the trivial always-reject baseline ($59.5\%$), and that even after recalibration only Direct Rating exceeds $\kappa = 0.1$: at this base rate none is usable as an autonomous gate, at any threshold. Replacing our score-based label with the recorded ICLR accept/reject decisions leaves the ordering unchanged, with Direct again ahead after recalibration and the always-reject baseline again essentially unbeaten (Appendix~\ref{app:decisionlabels}).

\section{RQ2: Model Dependence}
\label{sec:crossmodel}

\begin{table}[htbp]
\centering
\small
\setlength{\tabcolsep}{4pt}
\begin{tabular}{@{}lcccc@{}}
\toprule
\textbf{Model} & \multicolumn{2}{c}{\textbf{Best $\rho$}} & \multicolumn{2}{c}{\textbf{Best $\kappa$}} \\
\cmidrule(lr){2-3}\cmidrule(lr){4-5}
 & Paradigm & $\rho$ & Paradigm & $\kappa$ \\
\midrule
GPT-4o-mini   & Direct   & \textbf{0.422} & Elo      & \textbf{0.267} \\
Llama 3.3-70B & Swiss    & 0.140 & Swiss    & 0.060 \\
GPT-oss-120B  & Group & 0.338 & Group & 0.161 \\
\bottomrule
\end{tabular}
\caption{Best paradigm per model by ranking potential ($\rho$) and by decision utility, measured here by $\kappa$, abstract-only, 232-paper sample. The best paradigm differs by model and by criterion. Appendix~\ref{app:crossmodel_full} reports all fifteen configurations.}
\label{tab:crossmodel_summary}
\end{table}

\begin{figure*}[t]
\centering
\includegraphics[width=\textwidth]{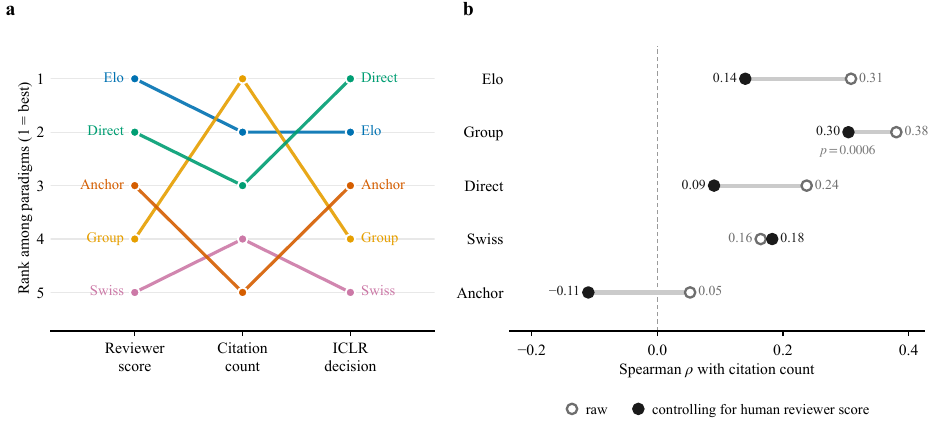}
\caption{External validity on the 126 test papers carrying a retrievable citation count, a complete set of paradigm scores and a recorded decision. \textbf{a}, Paradigm rank under three evaluation targets; the lines cross, so none is best on all three. The decision target counts the 18 withdrawn submissions as non-accepted, which leaves the ordering unchanged. \textbf{b}, Spearman $\rho$ with citation count, as an open circle (raw) joined to a filled circle (controlling for the mean reviewer score). Only Group Rating retains its correlation after the control ($\rho = 0.304$, $p = 0.0006$, surviving Bonferroni correction across the eighteen tests in Table~\ref{tab:citation}). Anchor's partial value is negative and Swiss's exceeds its raw value, so the two are given as positions on a common axis rather than as a decomposition.}
\label{fig:rq3}
\end{figure*}

To test whether the findings hold beyond GPT-4o-mini, we replicated all five paradigms (abstract-only) using Llama~3.3-70B and GPT-oss-120B; Table~\ref{tab:crossmodel_summary} gives the per-model best and Appendix~\ref{app:crossmodel_full} the full matrix.

These are fixed-threshold figures, so the $\kappa$ column carries the caveat of Section~\ref{sec:recalibration}. Recomputing every cell at its own base-rate-matched cut preserves the conclusion (Appendix~\ref{app:crossmodel_full}). Four findings follow. First, model choice matters at least as much as paradigm choice, as it does for LLM-generated preference distributions, which depend more on the choice of model than on the wording of the prompt \citep{huang2026preference}. Averaged over paradigms, the spread in $\rho$ across the three models is 0.293, slightly larger than the average spread across the five paradigms within a model (0.267); the margin is thin, so the more robust statement is that no single paradigm performs well on all three models, whereas GPT-4o-mini is the only model on which any paradigm exceeds $\rho = 0.35$. Second, Elo consistently reduces grade inflation relative to Direct for GPT-4o-mini ($+0.67$ vs.\ $+2.13$) and Llama~3.3-70B ($+0.19$ vs.\ $+3.42$), and the two comparative paradigms attain the lowest inflation for GPT-4o-mini (Swiss $+0.04$, Elo $+0.67$) and for Llama~3.3-70B (Elo $+0.19$, Swiss $+0.33$). This does not extend to GPT-oss-120B, where Elo inflation slightly exceeds Direct ($+1.11$ vs.\ $+0.99$) and Swiss shows the largest inflation of any configuration ($+3.82$); inflation reduction therefore appears to be a tendency of comparative scoring rather than a guaranteed property of it. Third, decision utility varies by model, from $\kappa = 0.267$ for GPT-4o-mini Elo down to $\kappa = 0.060$ for Llama~3.3-70B Swiss; weighted F1 can mislead here, since Llama~3.3-70B Elo reports weighted F1 $= 0.518$ despite $\kappa = -0.013$. Fourth, the best paradigm is model-dependent (Table~\ref{tab:crossmodel_summary}), and the separation between ranking potential and decision utility recurs across all fifteen configurations: GPT-4o-mini Direct has the highest $\rho$ but near-zero $\kappa$, while Elo converts comparable $\rho$ into $\kappa = 0.267$. For GPT-4o-mini Elo, pairwise ordering accuracy, the fraction of paper pairs ordered in the same direction as ground truth, is 64.5\% (vs.\ 50\% chance) and exceeds 75\% for paper pairs differing by at least 4 ground-truth points. Per-model score distributions and a $\rho$-versus-$\kappa$ view appear in Appendix~\ref{app:crossmodel}.

\section{RQ3: External Validity}
\label{sec:citation}

Do predicted scores track research impact? We correlate them with citation counts retrieved from Semantic Scholar \citep{kinney2023semanticscholar} in August 2026. Of the 232 test papers 163 matched a record, and because the dated-citation query is keyed on a DOI only the 135 carrying one were screened; 126 of those have at least one citation together with a complete set of paradigm scores and are the basis of every correlation and figure below. A further 26 matched, cited and fully scored papers are excluded for lacking a DOI, so the sample is enriched for work that was both disseminated and formally indexed. Because the source dates each citing paper, a 12-month and a 17-month window follow from one collection; they rank papers at $\rho = 0.992$ with no correlation moving by more than 0.010, so we report the 17-month figures and give the comparison in Appendix~\ref{app:citation}.

Human review scores correlate with citations at $\rho = 0.568$ ($p < 0.001$), well above every LLM paradigm; Group Rating is strongest at 0.380, followed by Elo at 0.308 and Direct at 0.238. Against reviewer scores on these same 126 papers Elo also leads Direct (0.353 against 0.293), reversing their order on the full 232-paper sample (Table~\ref{tab:paradigm}): the ranking between them is not stable across samples or targets.

The stratified columns of Table~\ref{tab:citation} qualify this, and the pattern is not specific to Elo. Splitting by ICLR decision, the human correlation holds within both strata (0.430 for accepted papers, 0.408 for rejected), so reviewers order papers by eventual impact within a pool and not merely across the accept/reject boundary; for ICLR 2020, \citet{tran2020openreview} found much weaker correlations between review score and citation rate within accepted posters and within rejected or withdrawn papers (0.17 and 0.22), and for 8{,}586 accepted ICLR 2017--2025 papers the mean review score is uncorrelated with a citation-embedding measure of disruption ($\rho = -0.003$; \citealp{huang2026catalyst}). Four of the five paradigms fail to reach significance among accepted papers, spanning $-0.082$ to $0.101$, while three are significant among rejected. Group Rating is the exception, at 0.330 and 0.420. For those four, the association with citations is carried by separating likely accepts from likely rejects rather than by ranking within the accepted pool.

A correlation with citations need not show that a paradigm contributes anything, since it may re-derive the human score. We therefore compute each paradigm's partial Spearman correlation with citations, controlling for the mean reviewer score. Most does not survive (Figure~\ref{fig:rq3}b): Elo falls from 0.308 to a non-significant 0.140 and Direct from 0.238 to 0.090. Group Rating behaves differently. It agrees with human scores at only $\rho = 0.242$ yet correlates 0.380 with citations and retains $\rho = 0.304$ after the control (bootstrap 95\% CI $[+0.129, +0.460]$, $p = 0.0006$, surviving Bonferroni correction across the eighteen tests in Table~\ref{tab:citation}). Swiss shows a weaker effect in the same direction that does not survive correction. Scoring five papers together therefore appears to surface impact-relevant signal that a single reviewer score does not capture. Given $n = 126$ and a single venue, we offer this as a direction worth testing rather than an established effect.

Which paradigm looks best also depends on the target: ranking the five under reviewer scores, citations and the recorded ICLR decision gives three different orderings (Figure~\ref{fig:rq3}a), and none wins on all three. Paradigm conclusions are contingent on the evaluation target, and a single target does not settle them; in LLM reasoning evaluation, similarly, replacing string matching with an LLM judge cuts one method's measured lead over another on HotpotQA from 13.0 to 4.7 points \citep{huang2026not}.

\section{Discussion}
\label{sec:discussion}

The separation between $\rho$ and $\kappa$ has a narrower reading than it first invites. Rank correlation measures ordering; a decision also requires that scores sit in the right place relative to the threshold. Direct Rating orders papers competitively and places them badly, the comparative paradigms place them better, but we cannot credit the comparison mechanism itself: our comparative ratings are min--max mapped onto the 1--10 scale while the absolute ones are not, and one order-preserving recalibration gives Direct Rating the better decision quality at lower cost (Section~\ref{sec:recalibration}). The defensible conclusion is about measurement: applying one fixed threshold to methods with different score placements ranks them largely by placement, so a study reporting only $\rho$, or only $\kappa$ at an arbitrary cut, can be misread in either direction.

The other two questions point the same way. A paradigm cannot be recommended independently of the model running it or the target it is scored against, so single-model single-target results are weaker evidence than they appear. Four of the five paradigms also lose their association with citations among accepted papers while human review retains it, separating likely accepts from likely rejects without ranking within the accepted pool. Group Rating is the exception, the one paradigm here behaving as a ranker rather than a filter.

The advantage of abstract-only input is consistent with abstracts acting as concentrated, author-curated signals, though our summaries are lossy enough that we cannot separate this from a pipeline effect; it is also a surface that automated rewriting targets, since rewrites that raised AI reviewer scores recast abstracts in more confident language \citep{baumann2026stop}. For deployment the population-scale results bound what is available (Table~\ref{tab:population}): no configuration exceeds the accuracy of always rejecting, and only Direct Rating exceeds $\kappa = 0.1$ even after recalibration, so Appendix~\ref{app:recommendations} reports operating points for comparison rather than as a recommendation.

\section{Conclusion and Future Work}
\label{sec:conclusion}

We compared five LLM-based review paradigms across three models, three input representations, and a citation outcome. On RQ1, Elo and Direct Rating attain comparable ranking potential ($\rho = 0.410$ vs.\ $0.422$) but very different decision utility at a fixed threshold, where Direct Rating flags 4.3\% of below-threshold papers against Elo's 66.4\%; a recalibration control locates the cause in placement rather than ordering, and at its own base rate Direct Rating moves ahead ($\kappa = 0.112$ vs.\ $0.079$) while making 24\% as many API calls as Elo. Abstract-only input wins throughout, and no configuration beats rejecting every paper at the deployment base rate. On RQ2, model choice matters at least as much as paradigm choice. On RQ3, human review scores track citations at $\rho = 0.568$, above every paradigm, and hold within both decision strata. Group Rating is the one paradigm that also holds within both, and it retains $\rho = 0.304$ with citations after controlling for the human score.

The replication also isolates a reporting issue: Cohen's $\kappa$ is sensitive to the spread of ground-truth scores, not only to class prevalence, the same Elo configuration reaching $0.267$ on a stratified sample and $0.076$ on the population. Absolute agreement should therefore be reported on a deployment-representative sample at a calibrated threshold. This is a separate mechanism from the marginal-dependence paradoxes of \citet{feinstein1990kappa}: the marginals barely move between the two designs, and what changes is the spread of the underlying scores.

The practical upshot is narrow but firm: an LLM reviewer is at best an aid to triaging a submission pool rather than a judge of an individual paper, and any claim about the best paradigm is a claim about a particular model, threshold and target.

Three directions follow: when $k$-run averaging of a cheap absolute paradigm beats switching to a comparative one; where the comparison-density elbow lies; and cost-sensitive threshold selection rather than raising $\kappa$ directly.

\section*{Limitations}

\paragraph{The paradigm comparison is not fully controlled.}
Input representation, model, prompt framing and decoding temperature are held constant across paradigms, but three confounds remain. First, the two families are not scaled alike: absolute paradigms emit 1--10 scores directly, whereas Elo and Swiss ratings are min--max mapped onto that scale, which constrains their means toward the middle independently of judgment quality. Reported inflation is therefore not a like-for-like measurement across families, and the recalibration control in Section~\ref{sec:recalibration} should be read as the primary comparison. Second, the paradigms are not compute-matched at population scale: Direct uses 3 calls per paper against Elo's 12.5. On the 232-paper set we ran the control (Appendix~\ref{app:reliability}). Giving Direct Elo's budget of five calls and averaging raises $\rho$ from 0.396 to 0.482, above Elo's 0.410, and leaves base-rate-matched $\kappa$ at 0.296 against Elo's 0.225 at its own base-rate-matched cut, so at equal compute Direct leads on both. The population-scale equivalent, 12.5 runs, remains unrun. Third, we estimate Elo ratings with the standard sequential update rather than a maximum-likelihood Bradley--Terry fit on the same comparisons; the latter is order-independent and would be the stronger estimator \citep{liu2025amelo,epstein2026allocate}.

\paragraph{Ceiling and ground truth.}
No single-run paradigm exceeded a Spearman correlation of 0.43, and at population scale the best is 0.204, against attenuation-corrected ceilings set by reviewer disagreement, $\rho = 0.951$ on the stratified sample and $\rho = 0.815$ at population scale (Section~\ref{sec:dataset}), bounds that constrain any predictor of this target, a second human reviewer included. Read against the ceiling of its own sample, the best stratified single-run result reaches $44.4\%$ of what is attainable and the population result $25.1\%$; the gap reflects the noise in the target as well as the limits of the paradigms. Appendix~\ref{app:robust} varies the aggregation of reviewer scores, and Appendix~\ref{app:decisionlabels} substitutes the recorded ICLR decisions, but neither addresses the noise in the underlying reviewer signal itself.

\paragraph{Statistical scope.}
Swiss's correlation is not significant at this sample size, so its ranking should be read cautiously. We report fifteen configurations without correction for multiple comparisons. The citation analysis covers the 126 of 232 papers that matched a Semantic Scholar record, carried a DOI through which dated citations could be retrieved, had at least one citation and carried a complete set of paradigm scores; requiring a DOI excludes a further 26 papers that would otherwise qualify, a subset enriched for disseminated work; its stratified columns rest on 74 accepted and 34 rejected papers, so the within-stratum estimates are the least precise numbers we report.

\paragraph{Generalization.}
Results come from one venue in one year within machine learning; venues with different review cultures may behave differently. The population-scale run draws from the same ICLR 2025 pool as the 232-paper sample, so it is a superset replication rather than external validation, and it covers only GPT-4o-mini. The two open-weight models we tested were weaker overall, with only GPT-oss-120B Group ($\rho = 0.338$) approaching the GPT-4o-mini range; our design cannot separate a proprietary/open-weight distinction from differences in scale and post-training. The open-weight runs also used condensed prompts for four of the five paradigms (Section~\ref{sec:methodology}), so for those paradigms prompt wording is confounded with model.

\paragraph{Fairness audit.}
Our Ethical Considerations section calls for auditing across research areas and author groups, and we did not run such an audit. This matters here because the best-performing configuration reads only the abstract, a surface that automated rewriting targets, and because \citet{baumann2026stop} report that AI reviewer scores can be moved by stylistic revision. A per-area breakdown of inflation and agreement at population scale is the natural next step and is feasible with the data we already hold.

\section*{Ethical Considerations}

This study uses publicly available ICLR submission metadata and reviewer scores from OpenReview and citation counts from Semantic Scholar; no private or personally identifying data beyond publicly listed information was used. Our findings characterize the capabilities and failure modes of LLM-based review and are intended to inform research on review assistance, not to replace human judgment. We caution against deploying any of these paradigms as an autonomous accept/reject gate. At the deployment base rate none of them exceeds the accuracy of rejecting every paper under our score-based rule, and only Direct Rating does so under the recorded ICLR decisions, by 0.3 points; the best attains $\kappa = 0.112$ after recalibration and the rest fall below 0.1, and all misclassify a substantial fraction of borderline papers; used without human oversight they could also systematically disadvantage particular topics, methods, or author groups, a risk we did not audit. LLM-based review may also reflect biases present in model training data \citep{yin2025dif}. We recommend that such systems be used only as a complement to human review, with audits for fairness across research areas and author demographics before any operational use.

\makeatletter
\ifacl@anonymize\else
\section*{Acknowledgments}

We thank the maintainers of the open datasets and platforms this work depends on, in particular OpenReview and Semantic Scholar, and the developers of the open-weight models we evaluated, Llama~3.3-70B and GPT-oss-120B.

\fi
\makeatother

\section*{Declaration on AI Assistance}

This declaration concerns AI assistance to the authors; the language models evaluated in this paper (Section~\ref{sec:methodology}) are the object of study, not writing aids.

AI-based tools were used in the preparation of this manuscript: for language and copy-editing, including drafting and revision of prose under author direction; for verification, checking bibliography entries against authoritative catalog records and recomputing reported statistics from the stored experimental artifacts, which surfaced a small number of numerical and citation errors that we corrected; and, as a pre-submission rehearsal, for generating model-written critiques of our own draft, several of which prompted additional controls that we implemented, ran on our own data, and verified before adopting.

No reported result was generated by an AI system: the experimental design, implementation and analysis are the authors' own, every number reported here is produced by our own pipeline, and the authors reviewed and take full responsibility for all content.

\bibliography{references}

\appendix

\section{Prompt Templates}
\label{app:prompts}

All paradigms share the same reviewer framing. Braces denote substituted fields; \texttt{summary\_section} is empty in the abstract-only condition and otherwise carries the generated summary. Appendix~\ref{app:worked-example} gives one of these templates fully instantiated, together with the model's responses. To save space, the Direct and binary-comparison templates below run the bulleted list of criteria into one line, and in the binary-comparison template each paper block as well, and shorten the sample value \texttt{Your reasoning here} in the JSON format to \texttt{...}; Appendix~\ref{app:worked-example} reproduces both prompts exactly.

\paragraph{Direct Absolute Rating.}
\begin{quote}\small\ttfamily
You are an experienced reviewer for a top-tier AI conference (ICLR 2025).\\[2pt]
Paper Information:\\
Title: \{title\}\\
Abstract: \{abstract\}\\
\{summary\_section\}\\[2pt]
Please provide an overall score for this paper on a scale of 1-10, where:\\
- 1-2: Strong reject (fundamental flaws, no contribution)\\
- 3-4: Reject (significant issues, limited contribution)\\
- 5-6: Borderline (some merit but concerns)\\
- 7-8: Accept (solid contribution, minor issues)\\
- 9-10: Strong accept (exceptional contribution)\\[2pt]
Consider: Novelty and originality; Technical quality and soundness; Clarity and presentation; Significance and impact.\\[2pt]
Provide your reasoning in 2-3 sentences, then give the score in JSON format:\\
\{"reasoning": "...", "score": <1-10>\}
\end{quote}

\paragraph{Anchor Rating.}
Reframed around three topically similar ICLR 2024 reference papers shown with their average review scores, with the instruction to rate the paper ``using the reference papers as calibration points'', the criteria introduced as a comparison with those papers, and the 1--10 score bands given without their descriptions.

\paragraph{Group Rating.}
As Direct, but presenting \texttt{\{num\_papers\}} papers together with the instruction to ``provide an exact score (1-10) for EACH paper based on comparative evaluation'', returning a JSON object keyed by paper.

\paragraph{Elo and Swiss (binary comparison).}
\begin{quote}\small\ttfamily
You are an experienced reviewer for a top-tier AI conference (ICLR 2025).\\[2pt]
You need to compare two papers and decide which one is better overall.\\[2pt]
{}**Paper A:** Title: \{title\_a\} Abstract: \{abstract\_a\} \{summary\_a\}\\
{}**Paper B:** Title: \{title\_b\} Abstract: \{abstract\_b\} \{summary\_b\}\\[2pt]
Compare based on: Novelty and originality; Technical quality and soundness; Clarity and presentation; Significance and impact.\\[2pt]
Which paper is better?\\[2pt]
Provide brief reasoning (1-2 sentences), then your decision in JSON format:\\
\{"reasoning": "...", "winner": "A" or "B" or "tie"\}
\end{quote}

\paragraph{Summarization.}
Vanilla summarization requests a 200--300 word summary covering main contribution and novelty, methodology, key results, and potential impact, with the instruction ``Be objective and factual. Do not include your personal opinions.'' Objective summarization requests the same length under four explicit headings (Background, Method, Results, Limitations) with the instruction ``Be purely objective. Only describe what the paper presents, without subjective judgments.''

\section{Worked Example: Instantiated Prompts and Model Responses}
\label{app:worked-example}

Appendix~\ref{app:prompts} gives the templates; this section gives one complete instantiation of them, reproduced from the experimental records. Every call shown here used \texttt{gpt-4o-mini} at temperature 0.3 under the system message ``You are an experienced research paper reviewer.'', with a token limit of 300 for Direct Rating, 400 for Anchor Rating, 500 for Group Rating and 200 for the binary comparison.

The subject is submission \texttt{P49gSPmrvN}, an ICLR 2025 paper that was rejected and that all three of its reviewers scored 1. It is shown because all three absolute paradigms returned 7 for it in the original 232-paper runs, so a single case exhibits the inflation reported in Section~\ref{sec:results}. The text below is unaltered apart from the elisions marked \texttt{[abstract elided]}, which remove every abstract in the anchor, group and comparison blocks, including the subject paper's own abstract in the group block; that abstract is given in full in the Direct Rating prompt. Neither the raw responses nor the per-comparison verdicts for Elo and Swiss were retained in our records, only the resulting ratings and, for the population-scale run, the request payloads, so the binary comparison is shown as a prompt alone.

\paragraph{Direct Rating, prompt as sent.}
\begin{quote}\small\ttfamily
You are an experienced reviewer for a top-tier AI conference (ICLR 2025).\\[3pt]
Paper Information:\\
Title: Time-dependent Development of Scientific Discourse: A Novel Approach Using UMAP and Word Embeddings\\
Abstract: This study presents a method for visualizing the time-dependent development of a scientific discipline using UMAP (Uniform Manifold Approximation and Projection) and text embeddings. This study demonstrates how the evolution of research interests and topics in a specific field can be mapped over time by encoding the abstracts of scholarly articles into a high-dimensional space and then projecting them into a 3D space. This computational approach converts the history of discourse into a point-cloud that can be further studied as a manifold and as a time series, which leads to new insights into the dynamics of scholarly discourse and the emergence and disappearance of research themes.\\[3pt]
Please provide an overall score for this paper on a scale of 1-10, where:\\
- 1-2: Strong reject (fundamental flaws, no contribution)\\
- 3-4: Reject (significant issues, limited contribution)\\
- 5-6: Borderline (some merit but concerns)\\
- 7-8: Accept (solid contribution, minor issues)\\
- 9-10: Strong accept (exceptional contribution)\\[3pt]
Consider:\\
- Novelty and originality\\
- Technical quality and soundness\\
- Clarity and presentation\\
- Significance and impact\\[3pt]
Provide your reasoning in 2-3 sentences, then give the score in JSON format:\\[3pt]
\{"reasoning": "Your reasoning here", "score": <1-10>\}
\end{quote}

\paragraph{Direct Rating, response as returned.}
\begin{quote}\small\ttfamily
\{"reasoning": "The paper presents a novel approach to visualizing the evolution of scientific discourse using UMAP and text embeddings, which is a relevant and timely topic in the field of AI and data visualization. The methodology appears sound, and the potential insights into research dynamics could have significant implications for understanding scholarly communication. However, the clarity of the presentation and the depth of the analysis could be improved to enhance the overall impact of the findings.", "score": 7\}
\end{quote}

\paragraph{Anchor Rating, reference block as sent.}
The remainder of the prompt is described in Appendix~\ref{app:prompts}. The three retrieved ICLR 2024 anchors average 3.7, yet the model placed the subject paper above all of them.
\begin{quote}\small\ttfamily
Reference Paper 1:\\
Title: From Language to 3D Worlds: Adapting Language Models for Point Cloud Perception\\
Abstract: [abstract elided]\\
Average Review Score: 5.0/10\\
-{}-{}-\\
Reference Paper 2:\\
Title: Causal Impact Index: A Causal Formulation of Citations\\
Abstract: [abstract elided]\\
Average Review Score: 3.0/10\\
-{}-{}-\\
Reference Paper 3:\\
Title: UGradSL: Machine Unlearning Using Gradient-based Smoothed Label\\
Abstract: [abstract elided]\\
Average Review Score: 3.0/10\\
-{}-{}-
\end{quote}

\paragraph{Anchor Rating, response as returned.}
\begin{quote}\small\ttfamily
\{"reasoning": "The paper presents a novel approach to visualizing the evolution of scientific discourse using UMAP and word embeddings, which is a fresh perspective in the field of text analysis and visualization. The technical execution appears sound, and the potential implications for understanding research dynamics are significant. However, while the methodology is interesting, it may not reach the same level of innovation or impact as the best work seen in the reference papers, leading to a score that reflects solid but not exceptional contributions.", "score": 7\}
\end{quote}

\paragraph{Group Rating, paper list as sent.}
The subject paper is Paper~D. This block comes from the original score-ordered grouping (Appendix~\ref{app:paradigms}), which is why all five papers in it were rated 1.0 to 1.4 by human reviewers; in the three randomized-grouping runs behind the reported abstract-only Group figures, the same paper received 5, 4 and 5.
\begin{quote}\small\ttfamily
{}**Paper A:**\\
Title: NEMESIS \textbackslash{}\textbackslash{}  Jailbreaking LLMs with Chain of Thoughts Approach\\
Abstract: [abstract elided]\\
-{}-{}-\\
{}**Paper B:**\\
Title: Balancing Differential Discriminative Knowledge For Clothing-Irrelevant Lifelong Person Re-identification\\
Abstract: [abstract elided]\\
-{}-{}-\\
{}**Paper C:**\\
Title: Systematic Review of Large Language Models: Applications, Limitations, Practical Usages and Future Directions\\
Abstract: [abstract elided]\\
-{}-{}-\\
{}**Paper D:**\\
Title: Time-dependent Development of Scientific Discourse: A Novel Approach Using UMAP and Word Embeddings\\
Abstract: [abstract elided]\\
-{}-{}-\\
{}**Paper E:**\\
Title: KL DIVERGENCE OPTIMIZATION WITH ENTROPY- RATIO ESTIMATION FOR STOCHASTIC GFLOWNETS\\
Abstract: [abstract elided]\\
-{}-{}-
\end{quote}

\paragraph{Group Rating, response as returned.}
\begin{quote}\small\ttfamily
\{\\
  "reasoning": "Paper A presents a novel approach to jailbreaking LLMs, addressing a critical issue in AI security, which enhances its significance and originality. Paper B introduces an innovative framework for lifelong person re-identification, but its contribution is somewhat niche and may not have as broad an impact. Paper C, while comprehensive, lacks novelty as it primarily reviews existing literature without presenting new findings. Paper D offers a unique method for visualizing scientific discourse, showcasing technical soundness and potential impact, but it may not appeal to a wide audience. Paper E introduces a novel optimization approach for GFlowNets, demonstrating strong technical quality and potential significance in generative modeling.",\\
  "scores": \{\\
    "Paper A": 8,\\
    "Paper B": 6,\\
    "Paper C": 5,\\
    "Paper D": 7,\\
    "Paper E": 8\\
  \}\\
\}
\end{quote}

\paragraph{Elo and Swiss, prompt as sent.}
Taken from the stored batch request payload for the first comparison of the population-scale run (\texttt{custom\_id: elo\_\_0}).
\begin{quote}\small\ttfamily
You are an experienced reviewer for a top-tier AI conference (ICLR 2025).\\[3pt]
You need to compare two papers and decide which one is better overall.\\[3pt]
{}**Paper A:**\\
Title: Understand Clean Generalization and Robust Overfitting in Adversarial Training from Two Theoretical Views: Representation Complexity and Training Dynamics\\
Abstract: [abstract elided]\\[3pt]
{}**Paper B:**\\
Title: Quantifying Likeness: A Simple Machine Learning Approach to Identifying Copyright Infringement in (AI-Generated) Artwork\\
Abstract: [abstract elided]\\[3pt]
Compare based on:\\
- Novelty and originality\\
- Technical quality and soundness\\
- Clarity and presentation\\
- Significance and impact\\[3pt]
Which paper is better?\\[3pt]
Provide brief reasoning (1-2 sentences), then your decision in JSON format:\\[3pt]
\{"reasoning": "Your reasoning here", "winner": "A" or "B" or "tie"\}
\end{quote}

\section{PDF Processing and Content Cleaning}
\label{app:preprocessing}

Papers were processed through a multi-stage pipeline: (1) PDF download from OpenReview, (2) conversion to Markdown using pymupdf4llm with a 60-second timeout, and (3) fallback to abstract-only for failed conversions. The pipeline achieved a 94.4\% success rate (219/232 papers), with 12 timeouts (5.2\%) and 1 other failure (0.4\%). To give the LLM focused content, the cleaning step removes line counters, formatting artifacts, and figure and table captions, drops inline citations in numeric and author-year formats, removes references and appendix sections, and normalizes whitespace. Content exceeding the token budget was truncated to 30{,}000 characters. These steps remove material that introduces noise or bias without contributing directly to quality assessment.

\paragraph{Software.} Anchor Rating retrieves its three in-context anchors with scikit-learn's \texttt{cosine\_similarity} over \texttt{text-embedding-3-small} abstract embeddings. The reported correlations (Pearson, Spearman and Kendall), the Wilcoxon signed-rank tests of Section~\ref{sec:methodology} and the trimmed mean of Appendix~\ref{app:robust} were computed with SciPy; AUC-ROC, Cohen's $\kappa$, macro and weighted F1, and NDCG with linear relevance gains were computed with scikit-learn. Accuracy, mean squared error, score inflation and the bootstrap resampling were computed directly in NumPy. Three quantities come from our own implementations rather than a package: the partial rank correlations of Section~\ref{sec:citation}, the base-rate-matched $\kappa$ and macro F1 of Section~\ref{sec:recalibration}, and the intraclass correlations, namely ICC(1,1) and ICC(1,$k$) in Section~\ref{sec:dataset} from a one-way random-effects analysis of variance corrected for the unequal number of reviewers per paper, and ICC(2,1) in Appendix~\ref{app:reliability} in the two-way random, single-rater, absolute-agreement form of \citet{shrout1979intraclass}. Matplotlib and seaborn were used for figures only. All analysis ran under Python 3.10; package versions were not pinned at run time and so are not reported.

\section{Additional Paradigm Implementation Details}
\label{app:paradigms}

\paragraph{Group construction and prompt position.} Group Rating's groups were originally drawn from consecutive rows of a file ordered by score bin, which left each group nearly homogeneous in ground-truth quality (Spearman between group index and mean reviewer score $0.986$; within-group ground-truth standard deviation $0.257$ against $2.12$ overall). Asking the model to separate five near-equals depresses the paradigm rather than flattering it: re-running the abstract-only condition with random groups and randomized within-prompt order raises $\rho$ against ground truth from $0.155$ (the original grouping, re-run alongside them; the original run reported in Table~\ref{tab:crossmodel} gives $0.101$) to between $0.278$ and $0.343$ over three seeds. The abstract-only Group Rating figures reported in this paper come from that corrected run, averaged over the three seeds; the vanilla and objective Group cells and the cross-model Group rows retain the original grouping and were not re-run. Every other figure reported in this paper is a single run at temperature 0.3 unless the accompanying text or caption states otherwise, so averaged and single-run entries appearing side by side are not like-for-like on run-to-run variation. Taken singly the three corrected Group runs give Spearman $\rho = 0.295$, 0.278 and 0.343 against ground truth.

A prompt-position effect survives the correction. A paper's score correlates negatively with its position in the prompt in every condition we ran, including after positions are randomized ($-0.325$ under the original grouping, $-0.217$ to $-0.327$ after randomization, mean $-0.279$ over the seven runs), so papers listed earlier are scored higher irrespective of content, a primacy effect also seen in GPT-4o-mini and Llama~3.3-70B, which choose the first-listed answer option 79.0\% and 77.1\% of the time against a 36.8\% chance rate \citep{huang2026cogbias}. This is a property of the paradigm rather than of the grouping, and it bounds what Group Rating can resolve within a prompt.

\paragraph{Token budget for Group Rating under the open-weight models.} Group Rating returns one JSON object scoring five papers, and the 500-token budget used for GPT-4o-mini truncated that response for the open-weight models. GPT-oss-120B parsed 0 of 232 papers at 500 tokens and was re-run in full at 2{,}000; the 17 Llama~3.3-70B groups containing an unparsed score (47 of their 85 papers) were re-run on the same basis. Temperature and the remaining token budgets are unchanged from the GPT-4o-mini run, as is the Direct prompt; the Anchor, Group and binary-comparison prompts are the condensed versions described in Section~\ref{sec:methodology}.

On the 232-paper set, Elo uses 1{,}160 comparisons, 10 per paper on average, drawn by uniform random pairing fixed in advance, so the comparison schedule is independent of intermediate ratings; ratings are then updated sequentially in the order in which the comparisons were returned, and the per-comparison outcomes were not retained. Swiss uses $\lceil\log_2(232)\rceil = 8$ rounds, pairing by current standings from round 2 onward while avoiding repeat matchups. At population scale, Elo uses 25 comparisons per paper on average and Swiss uses $\lceil\log_2(11{,}513)\rceil = 14$ rounds. The Elo update factor is $K=32$ and final ratings are min--max normalized to the 1--10 scale. Anchor Rating retrieves the three nearest anchors by cosine similarity of abstract embeddings. The relatively weak Swiss result on the 232-paper set may stem from strategic pairing forming local clusters that do not generalize to a global ranking, from too few rounds at that scale, and from binary win/loss outcomes carrying less information than Elo's probabilistic updates.

Wilcoxon signed-rank tests on absolute prediction error under abstract-only input give, for Elo versus each alternative, Direct ($\Delta|\text{error}| = +0.606$, $p < 0.001$), Anchor ($+0.912$, $p < 0.001$), Group ($+0.541$, $p < 0.001$), and Swiss ($+0.308$, $p < 0.01$); positive values mean the alternative has the larger error.

\section{Test--Retest Reliability of Direct Rating}
\label{app:reliability}

This appendix measures a source of variation that the bootstrap intervals reported elsewhere do not cover: those intervals resample papers, and so quantify sampling error in the evaluated set rather than run-to-run variation under stochastic decoding at temperature 0.3.

\paragraph{Compute-matched Direct.} A single Direct rating carries substantial decoding noise (ICC(2,1) $= 0.417$), so the natural control for the cost argument is to spend Elo's budget on Direct instead. We ran five independent Direct passes over the 232-paper set, abstract-only, with the published prompt and settings, and averaged the first $k$ of them. The Direct rows use the 230 papers with all five runs and the Elo row the 232-paper run of Table~\ref{tab:paradigm}; each base-rate-matched cut is the threshold whose reject share is closest to the true rate:

\begin{center}\small
\begin{tabular}{@{}cccc@{}}
\toprule
Calls/paper & $\rho$ & $\kappa$ at $\geq 6$ & base-rate-matched $\kappa$ \\
\midrule
1 & 0.396 & 0.046 & 0.288 \\
2 & 0.460 & 0.046 & 0.349 \\
3 & 0.460 & 0.046 & 0.322 \\
4 & 0.446 & 0.052 & 0.251 \\
5 & \textbf{0.482} & 0.046 & 0.296 \\
\midrule
\textit{Elo at 5} & \textit{0.410} & \textit{0.267} & \textit{0.225} \\
\bottomrule
\end{tabular}
\end{center}

Two things follow. At equal compute Direct overtakes Elo on ranking, 0.482 against 0.410, so the cost argument of Section~\ref{sec:recalibration} does not depend on giving Direct less budget. And averaging does not repair reject blindness: $\kappa$ at the fixed $\geq 6$ cut stays between 0.046 and 0.052 at every $k$, because extra runs reduce decoding noise without moving where the scores sit. That is the placement-versus-ordering distinction of Section~\ref{sec:recalibration} appearing again, from the compute side.

\begin{table}[htbp]
\centering
\small
\begin{tabular}{@{}lc@{}}
\toprule
\textbf{Metric} & \textbf{Value} \\
\midrule
ICC(2,1) absolute agreement              & 0.417 \\
Mean pairwise Spearman $\rho$            & 0.363 \\
Mean per-paper standard deviation        & 0.227 \\
Spearman ($\overline{\text{Direct}}$, GT)& 0.204 \\
\bottomrule
\end{tabular}
\caption{Test--retest reliability of the GPT-4o-mini Direct paradigm at temperature 0.3 (three independent runs, $n = 11{,}513$ papers, abstract-only). Reliability bands per \citet{koo2016guideline}: $<0.5$ poor, $0.5$--$0.75$ moderate, $0.75$--$0.9$ good, $>0.9$ excellent.}
\label{tab:reliability}
\end{table}

Table~\ref{tab:reliability} reports the reliability statistics summarized in Section~\ref{sec:reliability}. The mean pairwise Spearman $\rho$ of 0.363 across the three runs shows that even the \emph{ranking} induced by a single Direct pass is only moderately reproducible, and the Spearman correlation of the three-run mean against ground truth (0.204) is the population-scale Direct figure reported in Table~\ref{tab:population}.

\section{Population-Scale Error Localization}
\label{app:errorbin}

\begin{figure*}[htbp]
\centering
\includegraphics[width=\textwidth]{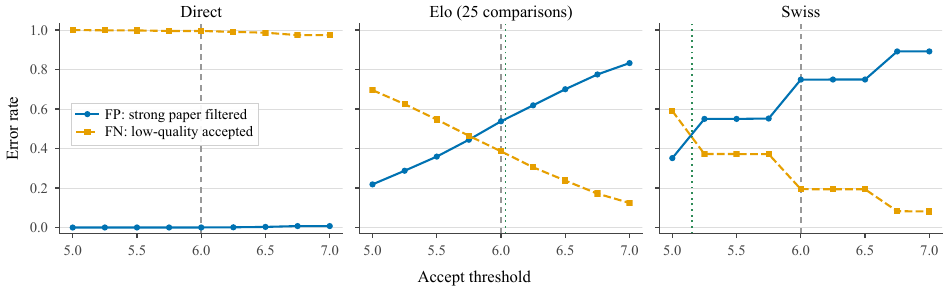}
\caption{Population-scale false-positive (strong paper wrongly filtered) and false-negative (low-quality paper wrongly accepted) rates as a function of the accept threshold, swept from 5 to 7. The dashed gray line marks the fixed $\geq 6$ threshold used in Table~\ref{tab:population}; the dotted green line marks the base-rate-matched threshold of Elo and Swiss; Direct Rating's, at 7.67, lies beyond the plotted range. Direct Rating's curves are flat across the plotted range because only 1.8\% of its predictions fall below 7, which is why a fixed threshold at 6 makes it appear degenerate.}
\label{fig:threshold_sweep}
\end{figure*}

\begin{figure*}[t]
\centering
\includegraphics[width=\textwidth]{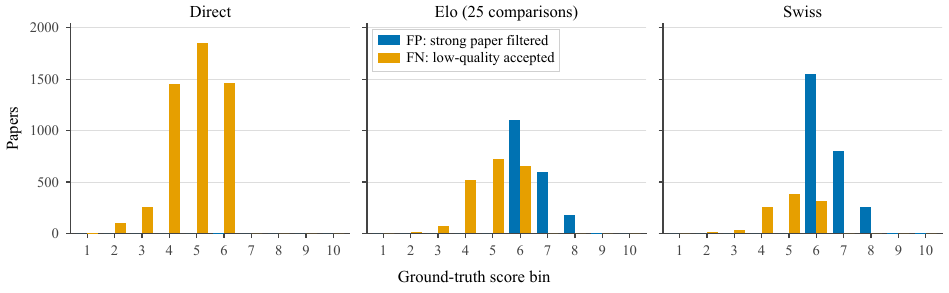}
\caption{Error localization by ground-truth score bin (population scale, threshold $\geq 6$). For each paradigm, blue bars show false-positive counts (strong papers wrongly filtered) and orange bars show false-negative counts (low-quality papers wrongly accepted). Direct's errors are almost entirely false negatives, since its inflation accepts nearly all low-quality papers, whereas Elo and Swiss distribute the error budget across both classes, with false positives concentrated at borderline ground-truth scores.}
\label{fig:error_by_bin}
\end{figure*}

Figure~\ref{fig:threshold_sweep} sweeps the accept threshold from 5 to 7, and Figure~\ref{fig:error_by_bin} localizes the population-scale errors by ground-truth score bin. Direct's errors are almost entirely false negatives, whereas Elo and Swiss distribute the error budget across both classes.

\section{Full Combination Analysis}
\label{app:full}

\begin{table}[htbp]
\centering
\small
\begin{tabular}{@{}lcccc@{}}
\toprule
\textbf{Input} & \textbf{$r$} & \textbf{$\rho$} & \textbf{$\tau$} & \textbf{Infl.} \\
\midrule
Abstract-Only & \textbf{0.333} & \textbf{0.334} & \textbf{0.254} & +1.46 \\
Objective & 0.193 & 0.180 & 0.137 & +1.75 \\
Vanilla & 0.177 & 0.172 & 0.132 & +1.72 \\
\bottomrule
\end{tabular}
\caption{Input representation compared, averaged across all five paradigms (GPT-4o-mini). $r$: Pearson; $\rho$: Spearman; $\tau$: Kendall; Infl.: grade inflation.}
\label{tab:summarization}
\end{table}

Table~\ref{tab:overall} reports all 15 paradigm-by-summarization combinations ranked by correlation.

\begin{table}[htbp]
\centering
\small
\setlength{\tabcolsep}{3pt}
\begin{tabular}{@{}llccccc@{}}
\toprule
\textbf{Paradigm} & \textbf{Input} & \textbf{Corr.} & \textbf{MSE} & \textbf{RMSE} & \textbf{Acc.} & \textbf{Infl.} \\
\midrule
Direct & Abstract & \textbf{0.405} & 8.42 & 2.90 & 0.134 & +2.13 \\
Group & Abstract & 0.405 & 7.87 & 2.81 & 0.121 & +1.98 \\
Elo & Abstract & 0.401 & \textbf{4.85} & \textbf{2.20} & 0.159 & +0.67 \\
Elo & Objective & 0.357 & 5.31 & 2.30 & 0.159 & +0.64 \\
Anchor & Abstract & 0.353 & 10.35 & 3.22 & 0.129 & +2.51 \\
Direct & Objective & 0.276 & 10.79 & 3.28 & 0.125 & +2.55 \\
Elo & Vanilla & 0.273 & 5.32 & 2.31 & \textbf{0.185} & +0.46 \\
Direct & Vanilla & 0.232 & 11.67 & 3.42 & 0.125 & +2.70 \\
Anchor & Vanilla & 0.193 & 12.23 & 3.50 & 0.125 & +2.78 \\
Anchor & Objective & 0.188 & 12.07 & 3.47 & 0.108 & +2.74 \\
Group & Vanilla & 0.116 & 10.96 & 3.31 & 0.113 & +2.34 \\
Swiss & Abstract & 0.100 & 6.71 & 2.59 & 0.147 & \textbf{+0.04} \\
Group & Objective & 0.078 & 10.48 & 3.24 & 0.104 & +2.18 \\
Swiss & Vanilla & 0.070 & 6.88 & 2.62 & 0.129 & +0.34 \\
Swiss & Objective & 0.066 & 7.51 & 2.74 & 0.138 & +0.64 \\
\bottomrule
\end{tabular}
\caption{All review methods, ranked by Pearson correlation (GPT-4o-mini, 232-paper sample). Input: input representation; Corr.: Pearson $r$ against ground truth; Acc.: exact-match rate on the rounded 1--10 score scale, which is not the binary accept/reject accuracy of Table~\ref{tab:classification}; Infl.: grade inflation. The Group vanilla and objective rows are computed on 230 papers, two having failed group assignment; the Group abstract row is the per-paper mean of the three randomized-grouping runs and covers all 232.}
\label{tab:overall}
\end{table}

\begin{figure*}[t]
\centering
\includegraphics[width=\textwidth]{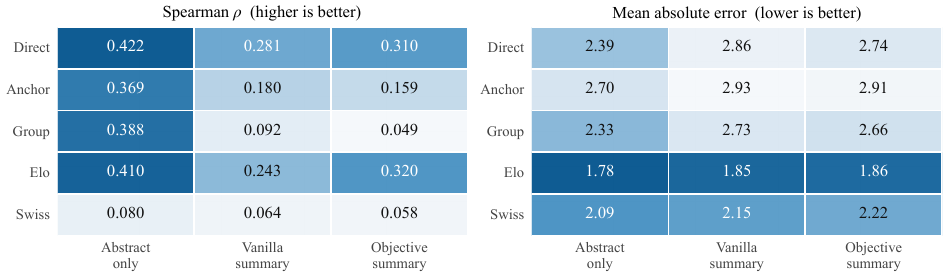}
\caption{Interaction between paradigm and input representation. Left: Spearman $\rho$ (higher is better). Right: Mean Absolute Error (lower is better). Abstract-only consistently outperforms across paradigms, and Elo achieves the lowest MAE regardless of input representation.}
\label{fig:sum_heatmap}
\end{figure*}

Within every paradigm, abstract-only yields the highest correlation. Averaged across input representations, Elo leads on Pearson correlation (0.344), followed by Direct (0.305), Anchor (0.245), Group (0.200), and Swiss (0.079). Bootstrap 95\% confidence intervals for Spearman $\rho$ under abstract-only evaluation are Direct [0.302, 0.537], Elo [0.302, 0.509], Anchor [0.249, 0.474], Group [0.269, 0.496], and Swiss [$-$0.049, 0.211]; the interval for Swiss includes zero. Figure~\ref{fig:sum_heatmap} shows the paradigm-by-representation interaction.

\section{Score Distributions by Input Representation and Paradigm}
\label{app:summdist}

\begin{figure*}[t]
\centering
\includegraphics[width=\textwidth]{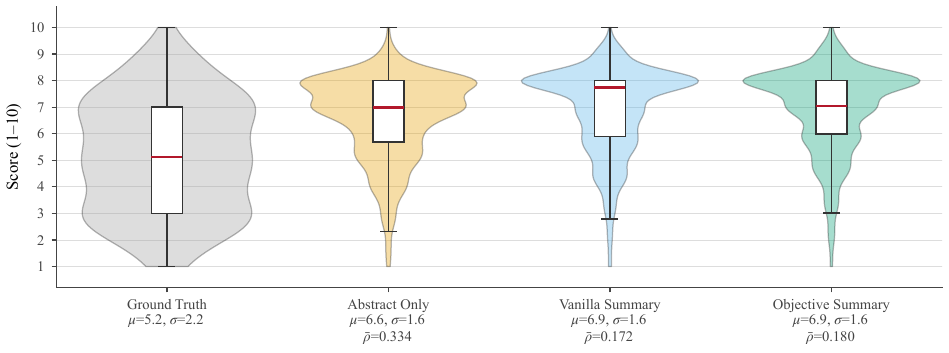}
\caption{Score distributions by input representation (all paradigms pooled, GPT-4o-mini). Violin shapes show density and box plots show quartiles; ground truth (gray) is shown for reference. The three representations differ little in spread (pooled $\sigma$ 1.56, 1.62 and 1.59 for abstract-only, vanilla and objective), and all are far narrower than ground truth ($\sigma = 2.16$). Annotated below each group: $\mu$ (mean), $\sigma$ (standard deviation), $\bar{\rho}$ (mean Spearman correlation with ground truth across paradigms).}
\label{fig:sum_distributions}
\end{figure*}

\begin{figure*}[t]
\centering
\includegraphics[width=\textwidth]{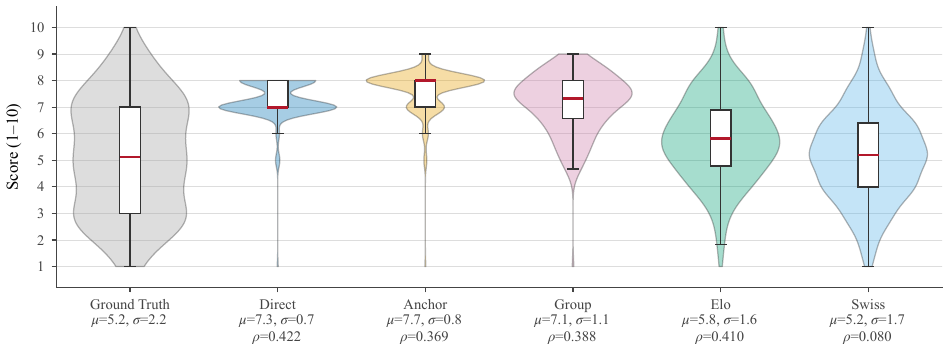}
\caption{Score distributions for the five paradigms vs.\ ground truth (abstract-only, GPT-4o-mini). Absolute methods (Direct, Anchor) compress scores into a narrow 6--8 range, while comparative methods (Elo, Swiss) preserve broader distributions closer to ground truth. Annotated below each: $\mu$ (mean), $\sigma$ (standard deviation), $\rho$ (Spearman correlation).}
\label{fig:paradigm_dist}
\end{figure*}

Figures~\ref{fig:sum_distributions} and \ref{fig:paradigm_dist} show the score distributions underlying the compression effect discussed in Sections~\ref{sec:summarization} and \ref{sec:classification}: absolute paradigms concentrate their predictions in a narrow high band, whereas comparative paradigms retain a spread close to ground truth.

\section{Stratified-Sample Classification Metrics}
\label{app:classification}

\begin{table}[htbp]
\centering
\small
\begin{tabular}{@{}lcccc@{}}
\toprule
\textbf{Method} & \textbf{Acc.} & \textbf{AUC} & \textbf{$\kappa$} & \textbf{Wgt.\ F1} \\
\midrule
Elo + Abstract & \textbf{64.2} & 0.672 & \textbf{0.267} & \textbf{0.645} \\
Elo + Objective & 58.6 & 0.645 & 0.163 & 0.590 \\
Elo + Vanilla & 58.2 & 0.619 & 0.134 & 0.584 \\
Direct + Abstract & 42.2 & \textbf{0.682} & 0.034 & 0.279 \\
Anchor + Abstract & 41.8 & 0.640 & 0.029 & 0.270 \\
Swiss + Abstract & 56.5 & 0.544 & 0.061 & 0.555 \\
\midrule
\textit{Random baseline} & \textit{50.0} & \textit{0.500} & \textit{0.000} & \textit{0.505} \\
\bottomrule
\end{tabular}
\caption{Binary classification performance (reject/accept) on the stratified 232-paper sample. Accuracy in \%; AUC is AUC-ROC on the continuous scores.}
\label{tab:classification}
\end{table}

\section{Ordering Quality and Error Analysis}
\label{app:ranking}

Table~\ref{tab:ranking} reports Normalized Discounted Cumulative Gain (NDCG) and top-$k$ precision, where top-$k$ precision is the fraction of a method's top-$k$ predicted papers that are truly above the accept threshold.

\begin{table}[htbp]
\centering
\small
\setlength{\tabcolsep}{4.5pt}
\begin{tabular}{@{}lcccc@{}}
\toprule
\textbf{Method} & \textbf{@10} & \textbf{@20} & \textbf{all} & \textbf{Top-20 Prec.} \\
\midrule
Elo + Objective & \textbf{0.752} & \textbf{0.757} & \textbf{0.933} & \textbf{0.700} \\
Elo + Abstract & 0.683 & 0.693 & 0.930 & 0.550 \\
Anchor + Abstract & 0.702 & 0.688 & 0.922 & 0.350 \\
Direct + Abstract & 0.669 & 0.691 & 0.926 & 0.000 \\
Swiss + Abstract & 0.662 & 0.650 & 0.906 & 0.300 \\
Direct + Objective & 0.665 & 0.661 & 0.918 & 0.050 \\
\bottomrule
\end{tabular}
\caption{Ordering-quality metrics for six representative methods, chosen to span the paradigms and input representations discussed in the text rather than to rank the top six on any single column. The first three numeric columns are NDCG@10, NDCG@20, and NDCG over all papers. Top-$k$ precision is computed with a stable sort, so for paradigms that assign the same score to large blocks of papers it reports one arbitrary tie-break rather than a property of the ordering; see the discussion of Direct + Abstract below.}
\label{tab:ranking}
\end{table}

All top methods achieve high overall NDCG ($>0.90$), but top-$k$ precision differs sharply. Elo + Objective identifies 70\% true accepts among its top-20 predictions. The corresponding figure for Direct + Abstract should not be read as a property of that paradigm: Direct emits only five distinct values on this sample and 87 of the 232 papers receive its highest score, so any top-20 is selected by the tie-break rule rather than by the model. That top block is $60.9\%$ true accepts against a $39.7\%$ base rate, so Direct's ordering is informative between blocks and simply unable to resolve within one; the 0.000 entry in Table~\ref{tab:ranking} is what the stable sort used here returns, and a random tie-break averages $61\%$ over 2{,}000 draws. A residual analysis shows that Direct, Anchor, and Group have systematic positive residuals (consistent over-prediction), Elo produces smaller residuals centered near zero, and Swiss has a near-zero mean residual but higher variance. All methods struggle at the extreme 1--2 and 9--10 ranges, reflecting both the rarity of such papers and the difficulty of identifying exceptional or fundamentally flawed work.

\section{Swiss Resolution and Elo Comparison Density}
\label{app:resolution}

\paragraph{Swiss resolution.} A Swiss tournament of $R$ rounds scores papers by win/tie/loss record, so it can express at most $2R+1$ distinct values, 29 at $R = \lceil\log_2 11{,}513\rceil = 14$. The run realized 23 of them across 11{,}513 papers, with clumps averaging 501 papers and the largest holding 2{,}404 (20.9\%). The consequence is a ceiling that no improvement in judgment quality can lift: $14.8\%$ of all paper pairs receive identical scores and cannot be ordered, and any threshold placed at the base rate lands inside a clump of 1{,}819 papers, 21.0\% of the 8{,}655 papers carrying a decision, containing both 1{,}123 true rejects and 696 true accepts, all resolved by tie-break. Elo's population-scale ratings are all distinct, so matching that resolution under the same $2R+1$ model would require 5{,}756 rounds, 411 times the 14 rounds and the API cost we report.

\paragraph{Why anchoring does not help.} The anchor set was selected for zero inter-reviewer variance, which collapses its score range: standard deviation 1.66 around a mean of 5.14 with a maximum of 8.0, against 2.16 and a full 1--10 range in the test set, and only 13 of 100 anchors below 3 or at or above 8. Averaging the three anchors retrieved per paper narrows the effective calibration signal further, to a standard deviation of 0.93. Anchor Rating is thus asked to calibrate a 1--10 judgment against exemplars that cluster near the middle and never reach the top of the scale. We tested the obvious alternative explanation, that the model is dragged toward whatever anchors it is shown, as LLM link predictions anchor on network scores supplied in the prompt ($\rho = 0.42$; \citealp{huang2026collab}), and found no support for it: predicted scores correlate at $\rho = -0.119$ with the mean score of the three anchors retrieved for that paper, so the failure is a lack of usable spread in the reference points rather than an anchoring bias.

\paragraph{Elo comparison density.} We ran the population-scale Elo at both 15 and 25 comparisons per paper. The 67\% increase in cost moves $\rho$ against ground truth from 0.132 to 0.142 while base-rate-matched $\kappa$ edges down from 0.084 to 0.079. The two density settings agree with each other at only $\rho = 0.811$, so a substantial share of the rating variance reflects which comparisons happened to be drawn rather than the underlying judgments. Density is therefore not the binding constraint at this scale: the extra comparisons buy a marginal gain in ranking and none in decision quality.

\section{Score Spread and Paradigm Combination}
\label{app:spread}

\paragraph{Why absolute agreement falls at population scale.} The stratified sample covers the 1--10 range far more evenly than the population: standard deviation 2.16 with $27.6\%$ of papers below 3 or at/above 8, against 1.16 with $61.4\%$ of papers in $[5,7)$. The accept rate is nearly identical in the two designs ($39.7\%$ against $40.5\%$), so the fall in $\kappa$ from 0.267 to 0.076 is a spread effect rather than a prevalence effect.

\paragraph{Score spread by paradigm.} Under abstract-only input, Direct and Anchor have $\sigma = 0.74$ and $0.81$ with 97\% and 94\% of predictions in the 6--8 band; Elo and Swiss reach $\sigma = 1.58$ and $1.67$ with 37\% and 28\% in band. Group Rating falls between them ($\sigma = 1.13$, 68\% in band): scoring five papers together spreads the scale even though the output remains absolute.

\paragraph{Combining paradigms.} Direct Rating and Elo agree with each other only weakly ($\rho = 0.345$ over all 11{,}513 scored papers), so their errors are partly independent. An equal-weight average of their $z$-scored outputs raises $\rho$ from 0.204 to 0.211 and recalibrated $\kappa$ from 0.112 to 0.115, but neither gain is distinguishable from zero under a paired bootstrap (95\% CI $[-0.004, +0.018]$ and $[-0.014, +0.020]$), so we do not claim a combination benefit.

\section{Ground-Truth Robustness}
\label{app:robust}

\begin{figure*}[t]
\centering
\includegraphics[width=\textwidth]{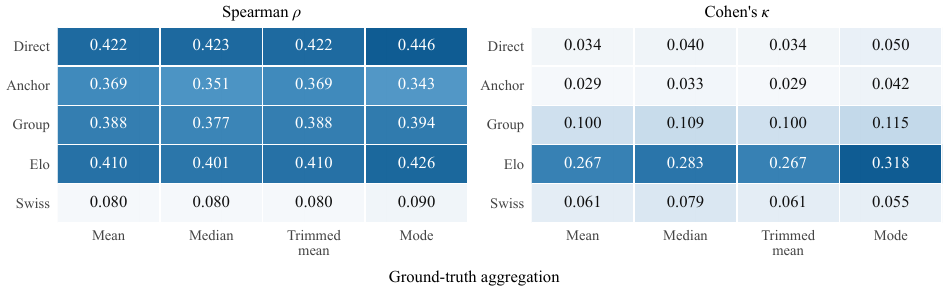}
\caption{Robustness check: Spearman $\rho$ (left) and Cohen's $\kappa$ (right) under four ground-truth aggregations, of which the trimmed mean coincides with the mean (Appendix~\ref{app:robust}). Paradigm rankings are identical across variants.}
\label{fig:robustness}
\end{figure*}

To check that the findings are not artifacts of the ground-truth aggregation, we recomputed the metrics using three distinct aggregations of reviewer scores: mean (our primary measure), median, and mode. A 10\%-trimmed mean was also computed but is not a fourth condition: papers here carry between two and seven reviewers, so a 10\% trim removes no scores and returns the mean exactly. The ranking-potential ordering is identical under all three, and under the corrected Group Rating run reported here it is Direct $>$ Elo $>$ Group $>$ Anchor $>$ Swiss. Elo maintains the highest decision utility across variants ($\kappa = 0.267$--$0.318$), well above Direct ($\kappa = 0.034$--$0.050$), so the separation between ranking potential and decision utility holds regardless of aggregation (Figure~\ref{fig:robustness}). These are fixed-threshold figures on the stratified sample; Section~\ref{sec:recalibration} shows that the separation itself is a placement effect, and this appendix establishes only that it is not an artifact of how reviewer scores are aggregated.

\section{Full Cross-Model Results}
\label{app:crossmodel_full}

Recomputing every cell of Table~\ref{tab:crossmodel} at its own base-rate-matched cut, rather than the fixed $\geq 6$ threshold, preserves the conclusion of Section~\ref{sec:crossmodel} and sharpens it. GPT-4o-mini Direct reaches $\kappa = 0.336$, above every other model-paradigm pair; Llama~3.3-70B peaks at 0.101 (Anchor) and GPT-oss-120B at 0.160 (Group). The best paradigm still differs by model, so none transfers at either threshold.

\begin{table*}[t]
\centering
\small
\begin{tabular}{@{}llcccccc@{}}
\toprule
\textbf{Model} & \textbf{Paradigm} & \textbf{Spearman $\rho$} & \textbf{Inflation} & \textbf{Cohen's $\kappa$} & \textbf{Macro F1} & \textbf{Wgt.\ F1} & \textbf{AUC-ROC} \\
 & & \scriptsize{(ranking)} & & \multicolumn{2}{c}{\scriptsize{(decision utility)}} & & \\
\midrule
\multirow{5}{*}{GPT-4o-mini}
  & Direct   & \textbf{0.422} & +2.13 & 0.034 & 0.330 & 0.279 & \textbf{0.682} \\
  & Anchor   & 0.369 & +2.51 & 0.029 & 0.323 & 0.270 & 0.640 \\
  & Group & 0.101 & +2.02 & 0.068 & 0.424 & 0.393 & 0.528 \\
  & Elo      & 0.410 & +0.67 & \textbf{0.267} & \textbf{0.633} & \textbf{0.645} & 0.672 \\
  & Swiss    & 0.080 & \textbf{+0.04} & 0.061 & 0.528 & 0.555 & 0.544 \\
\midrule
\multirow{5}{*}{Llama 3.3-70B}
  & Direct   & $-$0.004 & +3.42 & 0.006 & 0.292 & 0.234 & 0.509 \\
  & Anchor   & 0.079 & +3.03 & 0.000 & 0.303 & 0.249 & 0.537 \\
  & Group & 0.092 & +2.92 & 0.029 & 0.323 & 0.270 & 0.529 \\
  & Elo      & 0.042 & +0.19 & $-$0.013 & 0.493 & 0.518 & 0.516 \\
  & Swiss    & 0.140 & +0.33 & 0.060 & 0.530 & 0.553 & 0.541 \\
\midrule
\multirow{5}{*}{GPT-oss-120B}
  & Direct   & 0.194 & +0.99 & 0.082 & 0.476 & 0.458 & 0.613 \\
  & Anchor   & 0.024 & +0.89 & $-$0.005 & 0.459 & 0.453 & 0.507 \\
  & Group & 0.338 & +1.62 & 0.161 & 0.493 & 0.468 & 0.648 \\
  & Elo      & 0.035 & +1.11 & 0.007 & 0.393 & 0.362 & 0.499 \\
  & Swiss    & 0.068 & +3.82 & 0.011 & 0.300 & 0.244 & 0.514 \\
\bottomrule
\end{tabular}
\caption{Cross-model comparison, all fifteen configurations: five paradigms $\times$ three models (abstract-only, 232-paper sample). The Group rows, including GPT-4o-mini's, use the original score-ordered grouping (Appendix~\ref{app:paradigms}); the corrected GPT-4o-mini abstract-only Group run is reported in Table~\ref{tab:paradigm}. Three cells rest on fewer papers, where a response or a group assignment failed: GPT-4o-mini Group $n = 230$, GPT-oss-120B Direct $n = 230$ and GPT-oss-120B Anchor $n = 228$. Table~\ref{tab:crossmodel_summary} summarizes the per-model best.}
\label{tab:crossmodel}
\end{table*}

\section{Cross-Model Score Distributions and the \texorpdfstring{$\rho$--$\kappa$}{rho-kappa} View}
\label{app:crossmodel}

\begin{figure*}[t]
\centering
\includegraphics[width=\textwidth]{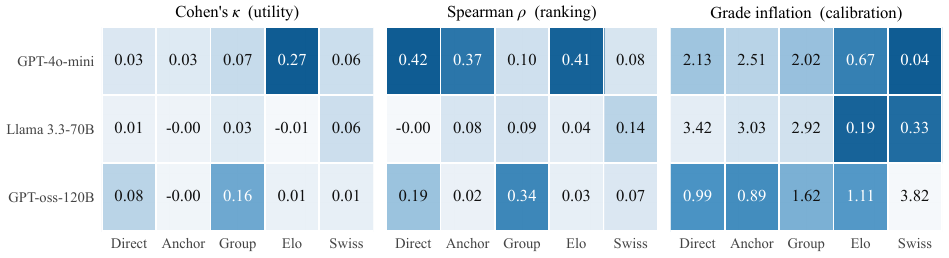}
\caption{Cross-model performance heatmaps. Left: Cohen's $\kappa$ (practical decision utility). Center: Spearman $\rho$ (ranking potential). Right: grade inflation (calibration). GPT-4o-mini Elo achieves the highest $\kappa$ despite slightly lower $\rho$ than Direct.}
\label{fig:crossmodel}
\end{figure*}

\begin{figure*}[t]
\centering
\includegraphics[width=0.93\textwidth]{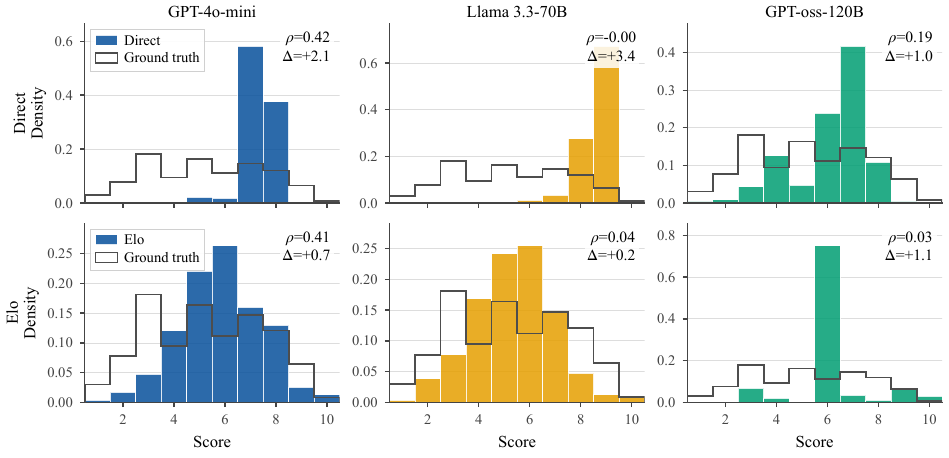}
\caption{Score distributions for Direct (top row) and Elo (bottom row); the columns are, left to right, GPT-4o-mini, Llama~3.3-70B, and GPT-oss-120B. GPT-4o-mini shows clear separation between ground truth and Direct predictions (inflation), which Elo corrects. Llama~3.3-70B Direct concentrates scores at 8--9, while its Elo output spreads scores without meaningful correlation. GPT-oss-120B shows moderate spread under Direct but collapses under Elo, with $75\%$ of papers in the single bin $[5.5, 6.5)$ and an interquartile range of 0.265, the narrowest panel here and a counterexample to the idea that comparative paradigms necessarily widen the scale.}
\label{fig:distributions_cross}
\end{figure*}

\begin{figure}[htbp]
\centering
\includegraphics[width=\columnwidth]{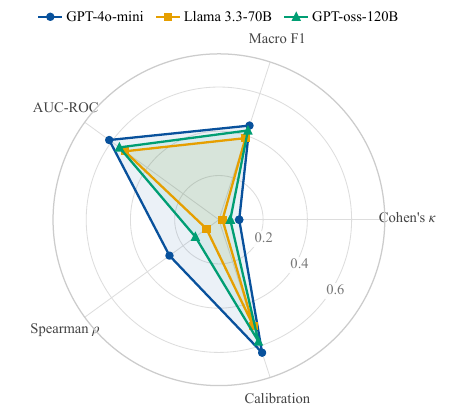}
\caption{Model capability profiles averaged across paradigms, on five axes: the decision utility metrics ($\kappa$ and macro F1), the threshold-free AUC-ROC, ranking potential ($\rho$), and a calibration axis plotting $1 - \text{inflation}/4$ so that a larger radius is better on every axis.}
\label{fig:radar}
\end{figure}

\begin{figure*}[t]
\centering
\includegraphics[width=\textwidth]{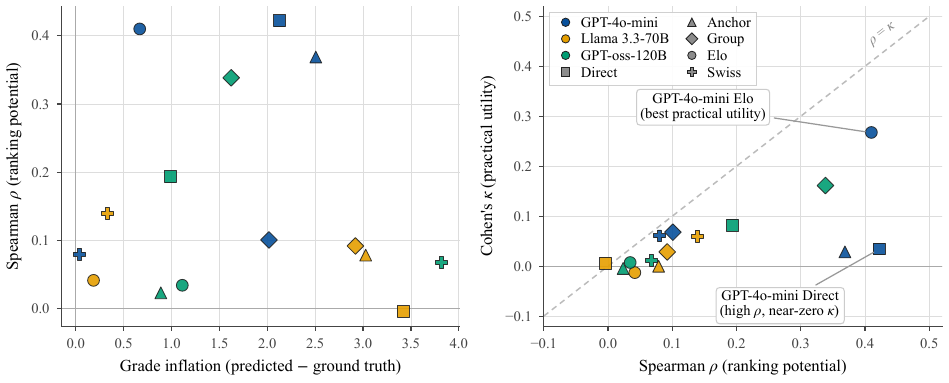}
\caption{Ranking potential versus grade inflation (left) and versus decision utility (right) for all 15 configurations. GPT-4o-mini Direct sits at high $\rho$ but near-zero $\kappa$, while Elo converts similar $\rho$ into higher $\kappa$ through calibration.}
\label{fig:tradeoff}
\end{figure*}

Figures~\ref{fig:crossmodel}--\ref{fig:tradeoff} give the per-model view behind Table~\ref{tab:crossmodel}, and make explicit that a high $\rho$ does not imply a high $\kappa$.

\section{Population-Scale Results}
\label{app:populationtab}

This appendix gives the population-scale analysis summarized in Section~\ref{sec:population} and used by the recalibration control of Section~\ref{sec:recalibration}. Table~\ref{tab:population} reports it at the fixed $\geq 6$ threshold and at each paradigm's base-rate-matched cut.

To test whether the result generalizes beyond the stratified sample, we re-ran the GPT-4o-mini abstract-only pipeline for Direct, Elo and Swiss on all 11{,}513 scored ICLR 2025 submissions with the same prompts and seed, taking the mean of three runs as Direct's score. Elo uses 25 comparisons per paper on average, of which the 57{,}565 added in the top-up from 15 ran with a 350-token output limit instead of 200: raising density from 15 moved $\rho$ against ground truth only from 0.132 to 0.142, indicating that the population-scale decline reflects the concentrated score distribution rather than an under-sampled comparison graph.

We exclude the 2{,}858 withdrawn papers, leaving 8{,}655 at a true reject rate of $59.5\%$, and take \textit{Reject} as the positive class, so a false positive is a strong paper wrongly filtered. Under this imbalance we read reject recall and macro F1 alongside accuracy \citep{saito2015precision} (Table~\ref{tab:population}).

Three observations follow. First, reject blindness sharpens at scale: Direct flags $0.5\%$ of the below-threshold papers here against $4.3\%$ on the stratified sample, with inflation essentially unchanged ($+2.20$ vs.\ $+2.13$).

Second, at this threshold the ordering of paradigms is preserved, but absolute agreement depends on the score distribution of the evaluated set. Elo retains the highest $\kappa$ (0.076), well below the $0.267$ of the 232-paper sample. This is not a class-prevalence effect, since the accept rate is nearly identical in the two designs; what changes is the spread of ground-truth scores, which is far wider in the stratified sample (Appendix~\ref{app:spread}). Both $\kappa$ and $\rho$ are raised by an abundance of easy, far-from-threshold papers and lowered when most papers sit near the boundary. We therefore recommend reporting absolute $\kappa$ on a \emph{deployment-representative} sample, whose score distribution matches the pool the system would actually screen, treating the stratified-sample $\kappa$ as a complementary diagnostic.

A structural limit explains much of Swiss's weakness: a tournament of $R$ rounds expresses at most $2R+1$ distinct scores, 29 at $R = 14$, and the run realized 23 clumps, $14.8\%$ of pairs are unorderable whatever the LLM judged, and the base-rate-matched cut falls inside a clump decided entirely by tie-break. Elo has no such ceiling but shows sharply diminishing returns to comparison density, with schedule noise rather than comparison count binding. Appendix~\ref{app:resolution} gives both analyses.

Third, the two comparative paradigms trade errors differently: Swiss reaches the highest reject recall (80.6\%) but wrongly filters $74.9\%$ of truly-accept papers, while Elo is more even (61.4\% and 53.8\%). Appendix~\ref{app:errorbin} gives the threshold sweep from which an operating point follows for a given cost ratio $c_{\text{FP}}/c_{\text{FN}}$, and shows the errors localizing in the borderline 4--7 band; population-scale cost is in Appendix~\ref{app:cost}.

\begin{table*}[t]
\centering
\small
\begin{tabular}{@{}llcccccc@{}}
\toprule
\textbf{Paradigm} & \textbf{Threshold} & \textbf{Spearman $\rho$} & \textbf{Inflation} & \textbf{Reject recall} & \textbf{Accuracy} & \textbf{$\kappa$} & \textbf{Macro F1} \\
\midrule
\multirow{2}{*}{Direct (mean of 3)} & fixed $\geq 6$    & \multirow{2}{*}{0.204} & \multirow{2}{*}{$+2.20$} & 0.5\%  & 40.8\% & 0.004 & 0.294 \\
                                    & base-rate matched &                        &                          & 66.9\% & 57.7\% & \textbf{0.112} & \textbf{0.555} \\
\midrule
\multirow{2}{*}{Elo (25 cmp/paper)} & fixed $\geq 6$    & \multirow{2}{*}{0.142} & \multirow{2}{*}{$+0.52$} & 61.4\% & 55.3\% & 0.076 & 0.538 \\
                                    & base-rate matched &                        &                          & 62.7\% & 55.6\% & 0.079 & 0.540 \\
\midrule
\multirow{2}{*}{Swiss (14 rounds)}  & fixed $\geq 6$    & \multirow{2}{*}{0.113} & \multirow{2}{*}{$+0.00$} & \textbf{80.6\%} & \textbf{58.1\%} & 0.061 & 0.511 \\
                                    & base-rate matched &                        &                          & 62.8\% & 55.6\% & 0.078 & 0.539 \\
\midrule
\textit{Always reject}              & ---               & \textit{---}           & \textit{---}             & \textit{100\%} & \textit{59.5\%} & \textit{0.000} & \textit{0.373} \\
\bottomrule
\end{tabular}
\caption{Population-scale results for GPT-4o-mini (abstract-only), at the fixed $\geq 6$ accept threshold and at each paradigm's own base-rate-matched threshold (the cut at which its predicted reject rate equals the true rate). Recalibration is order-preserving, so $\rho$ and inflation are unchanged by it. $\rho$ and inflation are computed on all 11{,}513 scored ICLR 2025 papers; decision metrics on the 8{,}655 non-withdrawn papers (true reject rate 59.5\%). Direct rows use the mean of three independent runs (ICC(2,1) $= 0.417$); because those scores are means of three integer ratings, Direct's closest achievable predicted reject rate is $62.4\%$ rather than $59.5\%$.}
\label{tab:population}
\end{table*}

\section{Results Under the Recorded ICLR Decisions}
\label{app:decisionlabels}

Section~\ref{sec:recalibration} defines Reject as a mean reviewer score below 6. The recorded ICLR 2025 decision is the operationally meaningful label, and it is available for all 8{,}655 non-withdrawn submissions. The two agree on $87.8\%$ of papers ($\kappa = 0.749$); the decision-based reject rate is $57.2\%$ against $59.5\%$ for the score rule. Table~\ref{tab:decisionlabels} repeats the population-scale analysis against the decisions.

\begin{table}[htbp]
\centering
\small
\setlength{\tabcolsep}{3.5pt}
\begin{tabular}{@{}llccc@{}}
\toprule
\textbf{Paradigm} & \textbf{Threshold} & \textbf{Rej.\ rec.} & \textbf{Acc.} & \textbf{$\kappa$} \\
\midrule
\multirow{2}{*}{Direct} & fixed $\geq 6$    & 0.5\%  & 43.0\% & 0.004 \\
                        & base-rate matched & 67.4\% & \textbf{57.5\%} & \textbf{0.118} \\
\midrule
\multirow{2}{*}{Elo}    & fixed $\geq 6$    & 62.3\% & 55.8\% & 0.093 \\
                        & base-rate matched & 61.3\% & 55.7\% & 0.095 \\
\midrule
\multirow{2}{*}{Swiss}  & fixed $\geq 6$    & 81.0\% & 57.2\% & 0.067 \\
                        & base-rate matched & 63.2\% & 55.5\% & 0.085 \\
\midrule
\textit{Always reject}  & ---               & \textit{100\%} & \textit{57.2\%} & \textit{0.000} \\
\bottomrule
\end{tabular}
\caption{Population-scale decision quality against the recorded ICLR 2025 accept/reject decisions rather than the $\geq 6$ score rule. The qualitative pattern is unchanged: Direct Rating is reject-blind at a fixed threshold and recovers under recalibration, where it moves ahead of Elo. Under these labels Direct Rating is the one configuration whose accuracy exceeds the always-reject baseline, and only by 0.3 points.}
\label{tab:decisionlabels}
\end{table}

\section{Citation Data Collection and Window Analysis}
\label{app:citation}

\begin{table}[htbp]
\centering
\small
\setlength{\tabcolsep}{4pt}
\begin{tabular}{@{}lccc@{}}
\toprule
 & \textbf{All} & \textbf{Accepted} & \textbf{Rejected} \\
\textbf{Score} & ($n=126$) & ($n=74$) & ($n=34$) \\
\midrule
Human review & \textbf{0.568}$^{\ast}$ & \textbf{0.430}$^{\ast}$ & 0.408$^{\ast}$ \\
Elo          & 0.308$^{\ast}$ & 0.101 & 0.410$^{\ast}$ \\
Group        & 0.380$^{\ast}$ & 0.330$^{\ast}$ & \textbf{0.420}$^{\ast}$ \\
Direct       & 0.238$^{\ast}$ & 0.084 & 0.120 \\
Swiss        & 0.165 & $-$0.019 & 0.399$^{\ast}$ \\
Anchor       & 0.052 & $-$0.082 & $-$0.033 \\
\bottomrule
\end{tabular}
\caption{Spearman $\rho$ between score and citation count, overall and within ICLR decision strata. $^{\ast}$ denotes $p < 0.05$. The human correlation survives stratification. As in Table~\ref{tab:paradigm}, the Group row is the per-paper mean of three randomized-grouping runs and the other rows are single runs. Group Rating is the only LLM paradigm to reach significance among accepted papers.}
\label{tab:citation}
\end{table}

\paragraph{Collection.} Citation counts were retrieved from Semantic Scholar in August 2026. Of the 232 test papers, 163 (70\%) matched a record, by DOI where one was available and by normalized-title similarity (Python \texttt{difflib} sequence ratio, accepted at $\geq 0.90$) otherwise. The window reconstruction retrieves each paper's citing papers by DOI, so the 28 matched papers without one could not be screened; of the remaining 135, the 126 with at least one citation and a complete set of paradigm scores are the basis of Table~\ref{tab:citation}. The 28 excluded papers are not uncited: 26 of them carry both citations and complete scores in the same August pull, and are omitted only because no DOI was available to query. The unmatched papers are mostly submissions that were neither accepted nor posted as preprints, so the analysis is restricted to disseminated work and the correlations are estimated under that restriction.

\paragraph{Window.} Rather than compare two separate collections, we retrieved every citing paper with its publication date, which allows both windows to be computed from one source and so separates the effect of the window from the effect of the source. Counting only citing papers published on or before 29 March 2026 gives the 12-month window; counting all of them gives the 17-month window. Across 9{,}821 citing papers, 4.0\% lacked a usable date and are excluded from the 12-month count. The two windows correlate at $\rho = 0.992$ with each other, and the score-citation correlations differ by at most 0.010 (human review 0.564 against 0.568; Elo 0.311 against 0.308; Group Rating 0.386 against 0.380; Direct 0.236 against 0.238).

\paragraph{Retrieving high-impact papers.} Ranking the 126 matched papers by each score and taking the top ten, human review recovers 5 of the 13 papers in the top citation decile against a random expectation of 1.0; Swiss recovers 4, Elo 2, and the remaining paradigms 1 or 2. These counts are not comparable across paradigms: with only 13 high-impact papers they are very noisy, and for Direct no paper scores strictly above the tenth-ranked value while for Group only four do, so their top ten is drawn entirely (Direct) or in six of ten places (Group) from a tie block and reflects the tie-break order rather than the scores. We report them for completeness only.

\paragraph{Relation to an earlier collection.} An earlier version of this analysis drew counts from a different source in March 2026 and reported much weaker correlations (human review $\rho = 0.189$, Elo $\rho = 0.201$), which we had read as parity between Elo and the human baseline. That reading does not survive re-collection. At the same 12-month cutoff, on the 123 papers present in both collections, the earlier data recorded $57\%$ as uncited with a median of 0, whereas the present collection records $8\%$ and a median of 11 at the same cutoff; $86\%$ of the papers previously recorded as uncited (60 of 70) have citations dated before it. The earlier figures reflected incomplete indexing of recent work rather than the absence of citations, and the apparent parity was an artifact of that sparsity.
\begin{table*}[t]
\centering
\small
\setlength{\tabcolsep}{4pt}
\begin{tabular}{@{}llccccl@{}}
\toprule
\textbf{Operating priority} & \textbf{Method} & \textbf{Calls/paper} & \textbf{Reject recall} & \textbf{FP rate} & \textbf{$\kappa$} & \textbf{When to use} \\
\midrule
Filter aggressively     & Swiss (14 rounds)  & 7    & \textbf{80.6\%} & 74.9\% & 0.061 & $c_{\text{FN}} \gg c_{\text{FP}}$ \\
Balanced quality filter & Elo (25 cmp/paper) & 12.5 & 61.4\% & 53.8\% & \textbf{0.076} & Best $\kappa$ at the fixed cut \\
Minimize false rejects  & Direct (mean of 3) & 3    & 0.5\%  & 0.0\%  & 0.004 & Equivalent to no filter \\
\bottomrule
\end{tabular}
\caption{Operating points at the fixed $\geq 6$ accept threshold, using population-scale numbers (Table~\ref{tab:population}). These are a comparison among the configurations tested, not a deployment recommendation: under base-rate-matched thresholding (Section~\ref{sec:recalibration}) Direct Rating attains the highest $\kappa$, and no configuration here reaches the $59.5\%$ accuracy of always rejecting. The cost ratio $c_{\text{FP}}/c_{\text{FN}}$ is the relative cost of filtering a strong paper (FP) versus admitting a low-quality one (FN).}
\label{tab:recommendations}
\end{table*}

\section{Deployment Cost and Operating Points}
\label{app:cost}

\label{app:recommendations}

The population-scale run used 259{,}035 API calls (Direct $3 \times 11{,}513$, Elo $143{,}912$ comparisons, and Swiss $80{,}584$ over 14 rounds) at a total cost of \$30.37 (Direct and Elo via the OpenAI Batch API at a 50\% discount; Swiss via real-time pricing), \$0.0026 per paper. Table~\ref{tab:recommendations} summarizes the operating points of the configurations tested at the fixed $\geq 6$ threshold; Section~\ref{sec:recalibration} gives the base-rate-matched figures, which reverse the ordering.

At the fixed $\geq 6$ threshold Direct is degenerate as a quality filter: its near-zero false-positive rate is achieved only by accepting essentially every paper (reject recall $0.5\%$, FN rate $99.5\%$), which is operationally equivalent to having no filter. Under base-rate-matched thresholding this reverses, and Direct attains the highest $\kappa$ of the three (Section~\ref{sec:recalibration}). Swiss attains high reject recall but wrongly filters $74.9\%$ of strong papers, so deploying it as a primary gate is defensible only when downstream human review of all Swiss-rejected papers is guaranteed, since the false-positive volume is large (2{,}626 strong papers wrongly filtered in the ICLR 2025 cohort). We therefore make no deployment recommendation: at the fixed threshold Elo offers the more even error split, under base-rate-matched thresholding Direct Rating attains the highest $\kappa$ with 24\% as many API calls as Elo, and none of the three reaches the accuracy of rejecting every paper.

\end{document}